\documentclass[journal]{IEEEtran}

\ifCLASSINFOpdf
\else
   \usepackage[dvips]{graphicx}
\fi
\usepackage{amsmath,amsfonts}
\usepackage{lineno,hyperref}
\usepackage{algorithmic}
\usepackage{array}
\usepackage[caption=false,font=normalsize,labelfont=sf,textfont=sf]{subfig}
\usepackage{textcomp}
\usepackage{stfloats}
\usepackage{url}
\usepackage{verbatim}
\usepackage{cite}
\usepackage{graphicx}
\def\BibTeX{{\rm B\kern-.05em{\sc i\kern-.025em b}\kern-.08em
    T\kern-.1667em\lower.7ex\hbox{E}\kern-.125emX}}
\usepackage{balance}
\usepackage{color}
\usepackage{float}

\usepackage{caption} 
\usepackage{amsfonts,amssymb}
\usepackage{amsmath}
\usepackage{algorithm}
\usepackage{booktabs}
\usepackage{multirow}

\begin{document}
\title{Enhanced Latent Multi-view Subspace Clustering}
\author{Long Shi, \IEEEmembership{Member, IEEE}, Lei Cao, Jun Wang, Badong Chen, \IEEEmembership{Senior Member, IEEE}  
\thanks{Long Shi and Lei Cao are with the School of Computing and Artificial Intelligence, and also with the Financial Intelligence and Financial Engineering
Key Laboratory of Sichuan Province, Southwestern University of Finance
and Economics, Chengdu 611130, China (e-mail: shilong@swufe.edu.cn, caolei2000@smail.swufe.edu.cn). These two authors contributed equally to this work.}

\thanks{Jun Wang is with the School of Management Science and Engineering,
Southwestern University of Finance and Economics, Chengdu 611130, China
(e-mail: wangjun1987@swufe.edu.cn).}

\thanks{Badong Chen is with the Institute of Artificial Intelligence andRobotics,Xi’an Jiaotong University, Xi’an 710049, China (e-mail: chenbd@mail.xjtu.edu.cn).}

\thanks{The work of Long Shi was partially supported by the National Natural Science Foundation of China under Grant 62201475 and Natural Science Foundation of Sichuan Province under Grant 2024NSFSC1436. The work of Badong Chen was supported by the National Natural Science Foundation of China under Grants U21A20485 and 61976175.}}


\maketitle

\begin{abstract}
Latent multi-view subspace clustering has been demonstrated to have desirable clustering performance. However, the original latent representation method vertically concatenates the data matrices from multiple views into a single matrix along the direction of dimensionality to recover the latent representation matrix, which may result in an incomplete information recovery. To fully recover the latent space representation, we in this paper propose an Enhanced Latent Multi-view Subspace Clustering (ELMSC) method. The ELMSC method involves constructing an augmented data matrix that enhances the representation of multi-view data. Specifically, we stack the data matrices from various views into the block-diagonal locations of the augmented matrix to exploit the complementary information. Meanwhile, the non-block-diagonal entries are composed based on the similarity between different views to capture the consistent information. In addition, we enforce a sparse regularization for the non-diagonal blocks of the augmented self-representation matrix to avoid redundant calculations of consistency information. Finally, a novel iterative algorithm based on the framework of Alternating Direction Method of Multipliers (ADMM) is developed to solve the optimization problem for ELMSC. Particularly, we theoretically analyze the convergence of ELMSC in detail. Extensive experiments on real-world datasets show that our proposed ELMSC is able to achieve higher clustering performance than some state-of-art multi-view clustering methods. Moreover, our experiments show that our method remains effective with randomly chosen parameters, demonstrating ELMSC’s practical potential.

\end{abstract}

\begin{IEEEkeywords}
ADMM, complementary information, consistent information, latent representation, multi-view subspace clustering, sparse regularization.
\end{IEEEkeywords}

\section{Introduction}

Subspace clustering for dealing with high-dimensional data has emerged as a research focus in the field of computer vision over the past decade, offering significant success in diverse scenarios including face clustering \cite{li2017structured, cao2023robust}, motion segmentation \cite{elhamifar2013sparse, huang2022multi, xing2020robust}, handwritten character recognition \cite{liu2012robust, lu2018subspace}. However, when confronted with insufficient or noisy data observations, subspace clustering inevitably experiences significant deterioration. This is because subspace clustering is essentially a single-view clustering technique, which is not tolerant of low-quality view data \cite{wu2021robust}. To address this challenge, multi-view clustering has arisen as a powerful solution that exploits abundant data sources from diverse domains \cite{xu2013survey, gao2015multi, lan2023multiview, yu2023sparse}. The concept of multi-view stems from intuitive observations made in various practical scenarios. For instance, in the context of web page products, each product page comprises textual descriptions, product images, and user engagement data. Unlike conventional subspace clustering, multi-view clustering methods can demonstrate impressive performance \cite{cao2015diversity, chao2017survey, qiang2021fast, jiang2022tensorial, chen2023joint}, even in the presence of limited-quality view data \cite{chao2019multi, chao2024incomplete, huang2021learning}.

Graph-based methods hold an important position in the field of multi-view clustering \cite{kumar2011co, li2015large, wang2019gmc, wong2019clustering}, primarily owing to the intrinsic capacity of graphs to capture and utilize relationships between data points across various views. In \cite{kumar2011co}, the authors introduced a co-regularization framework for multiview spectral clustering to align clustering hypotheses. \cite{zhan2018multiview} proposed a multiview consensus graph clustering method by defining an effective cost function for disagreement measure and imposing a rank constraint on the Laplacian matrix. \cite{zhu2018one} suggested integrating the shared affinity matrix learning stage and the \emph{k}-means clustering stage in a framework. To attain a parameter-free clustering model, \cite{nie2018multiview} developed a multi-view extension of spectral rotation with procrustes average. Multiple kernel learning has been acknowledged as another important technique for multi-view clustering. \cite{liu2016multiple} proposed a multiple kernel \emph{k}-means clustering method with a matrix-induced regularization to improve the diversity of the selected kernels. With late fusion alignment maximization, \cite{wang2019multi} proposed an effective clustering method to reduce the computational cost and simplify the optimization procedures. Furthermore, extensive research has been conducted in the field of deep multi-view learning. Huang \emph{et al.} introduced a method for multi-view spectral representation that applies an orthogonal constraint as a layer \cite{huang2021deep}. Pan \emph{et al.} incorporated the contrasting learning with multi-view graph \cite{pan2021multi}. Fan \emph{et al.} proposed a task-guided deep technique for attributed multi-view graph clustering \cite{fan2020one2multi}. Lu \emph{et al.} integrated high-order random walks with contrastive multi-view clustering to tackle the false negative issue \cite{lu2024decoupled}. For more advanced graph-based, multiple kernel and deep learning methods, readers are encouraged to refer \cite{zhou2019multiple, wang2020deep, li2021consensus, lin2021multi, li2022local, liu2022simplemkkm, chen2023multiview, yan2021deep}.

In addition to the previously mentioned methods, multi-view subspace clustering \cite{fang2017orthogonal, lan2022generalized} has also gained significant popularity. \cite{gao2015multi} performed subspace clustering on each view while ensuring the coherence of the clustering structure among different views. \cite{brbic2018multi} enforced the multi-view subspace clustering with low-rank and sparse structures. In order to comprehensively explore the underlying data distribution across distinct views, \cite{luo2018consistent} proposed a subspace learning method that simultaneously takes consistency and diversity into account. Wang \emph{et al.} investigated the Frobenius norm constraint and adjacency similarity learning to explore both global and local view information \cite{wang2022frobenius}. Given the advantages of capturing high-order correlations, tensor-based multi-view subspace clustering methods have been reported \cite{jia2021multi, wang2023robustness}. Moreover, considerable efforts have been dedicated to addressing the challenges of large-scale datasets \cite{qiang2021fast, yang2022efficient} and incomplete views \cite{liu2021novel, wen2023graph, yu2023sparse, yang2022robust}. Motivated by the underlying assumption that multiple views are generated from one shared latent representation, the latent representation-based multi-view subspace clustering methods have garnered much attention \cite{zhang2017latent, zhang2018generalized, chen2020multi}. In contrast to conventional multi-view subspace clustering methods, the latent representation method first recovers the latent space matrix, followed by the learning of the self-representation matrix based on the derived latent representation matrix. Despite the promising results achieved, the existing latent representation-based methods directly concatenate the data matrices from multiple views into a single matrix along the direction of dimensionality. This data concatenation strategy only takes into account the unique information from each view, and fails to exploit the similarity, also known as consistency, between different views, thereby limiting its ability to facilitate a comprehensive recovery of the latent representation matrix.

To this end, we propose a novel latent representation clustering method called Enhanced Latent Multi-view Subspace Clustering (ELMSC). We take into account the complementarity and consistency of various views simultaneously to construct an augmented multi-view data matrix, which is useful for learning an exhaustive latent representation matrix. For the purpose of concept alignment, we refer to the latent representation in this paper as the augmented latent representation matrix. Our main contributions include:

\begin{itemize}
    \item To guarantee the complementary information of each view as well as exploit the consistent information across views, we construct an augmented data matrix, in which the block-diagonal locations stack the data matrices from different views and the non-block-diagonal entries consist of the similarity information between different views. 
    \item We enforce a sparse regularization for the non-diagonal blocks of the augmented self-representation matrix to avoid redundant calculations of consistent information between different views. It is worth noting this regularization does not affect the diagonal blocks, ensuring the preservation of complete complementary information.
    \item An iterative optimization algorithm is developed to solve the optimization problem for ELMSC by using the framework of Alternating Direction Method of Multipliers (ADMM) \cite{boyd2011distributed}.
    \item We establish the relationship between the traditional latent multi-view technique and our method under some specific conditions, indicating that our approach can be regarded as an augmented variant of the traditional method. We also theoretically analyze the convergence of ELMSC. 
\end{itemize}

\noindent Experiments on six real-world datasets have demonstrated that the proposed ELMSC method outperforms some advanced multi-view clutsering methods. Moreover, we examine the performance of ELMSC when parameters are randomly selected. Experiments show that ELMSC with randomly selected parameters can achieve desirable performance that is only slightly inferior to ELMSC with optimally chosen parameters, or nearly matches the top-performing baseline method. This implies that ELMSC is not significantly sensitive to parameter selection, thereby addressing the difficulties of challenging parameter selection that many existing methods face in practical implementation.


\section{Preliminaries}

\subsection{Multi-view Subspace Clustering}

Multi-view Subspace Clustering (MVSC) \cite{gao2015multi} aims to address the limitations of traditional subspace clustering methods by combining information from diverse sources to reveal underlying structures and relationships. Consider multi-view data $\{\mathbf X^{(l)}\}^v_{l=1}$, where $\mathbf X^{(l)}\in \mathbb{R}^{d_l \times n}$ denotes the $l$-th view data with $d_l$ dimensions. The objective optimization problem of MVSC can be defined as
\begin{equation}
\begin{aligned}
\label{eq01}
&\min_{\mathbf{Z^{(l)},\mathbf Z}}\sum_l\lVert\mathbf X^{(l)}-\mathbf X^{(l)}\mathbf Z^{(l)}\rVert^2_F + \lambda f(\mathbf Z, \mathbf Z^{(l)}), 
\\
& s.t. \,\mathbf Z_l\geq 0, ({\mathbf Z^{(l)}})^T\mathbf 1 = \mathbf 1,  
\end{aligned}
\end{equation}
where $\lVert\cdot\rVert_F$ represents the $F$-norm, $\mathbf Z^{(l)}\in \mathbb{R}^{n\times n}$ denotes the non-negative self-representation matrix, $\mathbf Z$ accounts for the unified self-representation matrix, $\lambda$ is a balance parameter, and $f(\cdot)$ represents the unified regularization term. The equation $({\mathbf Z^{(l)}})^T\mathbf 1 = \mathbf 1$ enforces the constraint that the coefficients of data points within a subspace should sum up to $1$, ensuring that the data points are fully represented within their respective subspaces during the clustering process.     

After acquiring the unified self-representation matrix $\mathbf Z$, the subsequent procedure is to perform spectral clustering to obtain the ultimate clustering results. This process involves the construction of the affinity matrix via $\mathbf{W} = \frac{\lvert\mathbf{Z}\rvert + \lvert\mathbf{Z}^\intercal\rvert}{2}$, followed by the resolution of an optimization problem  
\begin{equation}
\label{eq02}
\min_{\mathbf F}{\rm Tr}(\mathbf F^T\mathbf L\mathbf F), \ s.t. \,\mathbf F^T\mathbf F =  \mathbf I,
\end{equation}
where ${\rm Tr}(\cdot)$ takes the trace of a matrix, $\mathbf L = \mathbf D-\mathbf W$ is the Laplacian matrix with diagonal matrix defined as $d_{ii} = \sum^n_{j=1} z_{ij}$ \cite{nie2016constrained}, and $\mathbf F$ is the cluster indicator matrix.  

\subsection{Latent Multi-view Subspace Clustering}

In Latent Multi-view Subspace Clustering (LMSC) \cite{zhang2017latent}, the self-representation matrix is learned from the recovered latent space representation, in contrast to the conventional multi-view subspace clustering methods that rely on data representations. The objective optimization of LMSC is formulated as
\begin{equation}
\begin{aligned}
\label{eq03}
&\min_{\mathbf P,\mathbf H,\mathbf Z}f_h(\mathbf X,\mathbf P\mathbf H) + \lambda_1 f_r(\mathbf H,\mathbf H\mathbf Z) + \lambda_2\Omega(\mathbf Z)\\
&{\rm with}\; \mathbf X = \left[
\begin{array}{c}
\mathbf X^{(1)} \\
\cdots \\
\mathbf X^{(v)}
\end{array}
\right]
{\rm and}\; \mathbf P = \left[
\begin{array}{c}
\mathbf P^{(1)} \\
\cdots \\
\mathbf P^{(v)}
\end{array}
\right],
\end{aligned}
\end{equation} 
where $\mathbf X\in \mathbb{R}^{d\times n}$ denotes the reconstruction data matrix with $d=\sum_l d_l$, $\mathbf P\in \mathbb{R}^{d\times k}$ is the projection matrix, $\mathbf H\in \mathbb{R}^{k\times n}$ represents the latent representation matrix, $\mathbf Z$ is the self-representation matrix also known as the reconstruction coefficient matrix, $\lambda_1$ and $\lambda_2$ are the balance parameters. The function $f_h(\mathbf X,\mathbf P\mathbf H)$ serves to recover the underlying latent representation $\mathbf H$, $f_r(\mathbf H,\mathbf H\mathbf Z)$ is used to learn the self-representation matrix $\mathbf Z$, and $\Omega(\mathbf Z)$ is a regularization term.   

\section{Proposed Methodology}

\begin{figure*}[t]
\centering
\includegraphics[width=0.95\textwidth]{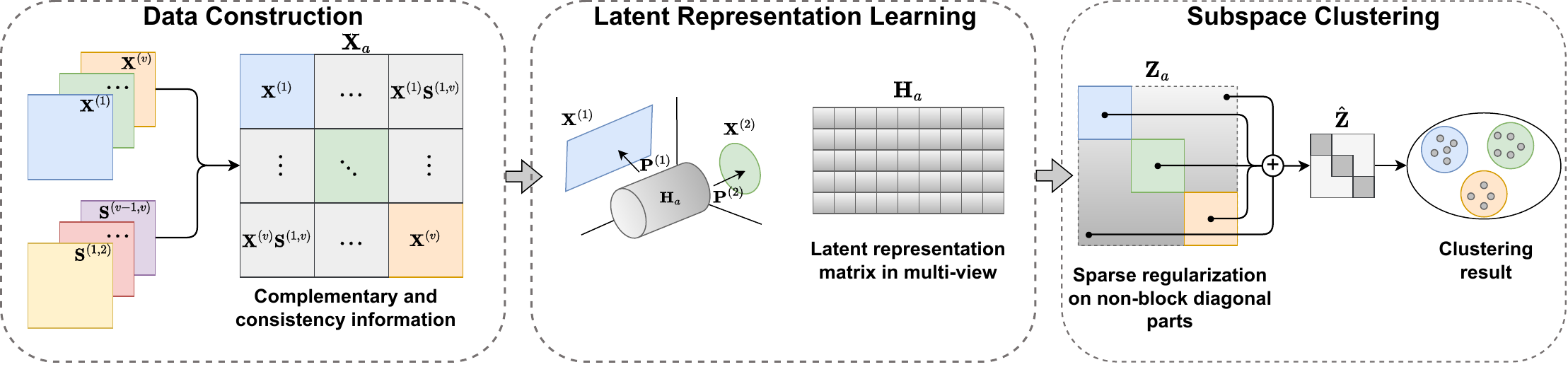} 
\caption{Framework of ELMSC.}
\label{Fig_fra}
\end{figure*}

In this section, we will comprehensively introduce the proposed ELMSC methodology, offering a detailed presentation of its motivation, formulation, optimization procedures, and an analysis of its computational complexity. The framework of ELMSC is shown in Fig. \ref{Fig_fra}.

\subsection{Motivation}

As shown in Eq. (\ref{eq03}), LMSC vertically concatenates the data matrix of each view to form a multi-view matrix for latent representation learning. While this construction method captures the inherent complementarity across diverse views, it lacks the capability to exploit the consistent information between views. Thus, there arises a need to devise a more effective multi-view data matrix construction strategy to enhance representation quality, thereby facilitating the recovery of an improved latent representation. The motivation and contribution of our paper will be further emphasized towards the end of the Problem Formulation subsection, through a comparison between the conventional data reconstruction method and our approach, as well as an interpretation of our data construction that relates with attention mechanism.

\subsection{Problem Formulation}

In order to enhance the multi-view data representation, we define an augmented data matrix $\mathbf X_a$, in which the block-diagonal locations stack the data matrices from different views to preserve the complementary information, while the non-block-diagonal regions contain inter-view similarities to capture the consistent information. Specifically, the augmented data  matrix $\mathbf X_a\in \mathbb{R}^{d\times vn}$ takes the form   
\begin{equation}
\label{eq04}
\mathbf X_a = \left[
\begin{array}{c c c c}
\mathbf X^{(1)}, & \mathbf X^{(1,2)}, & \cdots, & \mathbf X^{(1,v)} \\
\mathbf X^{(2,1)}, & \mathbf X^{(2)}, & \cdots, & \mathbf X^{(2,v)} \\
\vdots & \vdots & \vdots & \vdots\\
\mathbf X^{(v,1)}, & \mathbf X^{(v,2)}, & \cdots, & \mathbf X^{(v)}\\
\end{array}
\right],
\end{equation}
where $\mathbf X^{(p,q)}\in \mathbb{R}^{d_p\times n}$ denotes the correlation matrix between the $p$-th and $q$-th views that is calculated by 
\begin{equation}
\label{eq05}
\mathbf X^{(p,q)} = \mathbf X^{(p)}\mathbf S^{(p,q)}, \; {\rm for}\; p \neq q. 
\end{equation}
In Eq. (\ref{eq05}), $\mathbf S^{(p,q)}$ represents the cosine similarity matrix \cite{tan2023metric} with its element $S^{(p,q)}_{i,j}$ computed by 
\begin{equation}
\label{eq06}
    \mathbf{S}^{(p,q)}_{i,j} = \frac{\mathbf{\Tilde{X}}^{(p)}_{:,i}\cdot\mathbf{\Tilde{X}}^{(q)}_{:,j}}{2\Vert \mathbf{\Tilde{X}}^{(p)}_{:,i} \Vert_2 \Vert \mathbf{\Tilde{X}}^{(q)}_{:,j} \Vert_2}+\frac{1}{2},
\end{equation}
where $\mathbf{\Tilde{X}}^{(p)}$ and $\mathbf{\Tilde{X}}^{(q)}$ are the data matrices after dimensionality reduction via Principal Component Analysis (PCA) \cite{abdi2010principal, bro2014principal}, a procedure aimed at aligning data dimensions between the $p$-th and $q$-th views. Further details on the implementation of PCA will be illustrated in the section of Experiments. This alignment is essential in enabling the computation of cosine similarity. The symbols $\mathbf{\Tilde{X}}^{(p)}_{:,i}$ and $\mathbf{\Tilde{X}}^{(q)}_{:,j}$ denote the $i$-th row and the $j$-th row of $\mathbf{\Tilde{X}}^{(p)}$ and $\mathbf{\Tilde{X}}^{(q)}$, respectively. $\mathbf{\Tilde{X}}^{(p)}_{:,i}\cdot\mathbf{\Tilde{X}}^{(q)}_{:,j}$ calculates the inner product of two vectors. From Eq. (\ref{eq06}), it is evident to infer that $0\leq \mathbf S^{(p,q)}_{i,j}\leq 1$ and $\mathbf S^{(p,q)} = \mathbf S^{(q,p)}$. 

Subsequently, we focus on the formulation of the objective optimization problem. First, the objective function for learning the underlying latent representation is defined as follows
\begin{equation}
\label{eq07}
    \min_{\mathbf{P},\mathbf{H}_a} J_1\left(\mathbf{X}_a,\mathbf{P H}_a \right),
\end{equation}
where $\mathbf H_a\in \mathbb{R}^{k\times vn}$ denotes the augmented latent representation matrix, and $J_1(\cdot)$ is the loss function associated with the latent presentation. It is worth noting that in contrast to the objective function concerning latent representation in LMSC, as shown in Eq. (\ref{eq03}), the constructed multi-view data matrix $\mathbf X_a$ and the latent representation matrix $\mathbf H_a$ in our method undergo augmentation in the formulation, which is expected to offer enhanced representation capabilities.    

Then, with the latent representation matrix in Eq. (\ref{eq07}), the objective function of self-representation based subspace clustering is formulated as
\begin{equation}
\label{eq08}
    \min_{\mathbf{Z}_a} J_2\left(\mathbf{H}_a,\mathbf{H}_a\mathbf{Z}_a \right)+{\rm{Reg}}\left(\mathbf{Z}_a \right),
\end{equation}
where $\mathbf Z_a\in \mathbb{R}^{vn\times vn}$ denotes the augmented self-representation matrix, $J_2(\cdot)$ is the loss function associated with the data self-representation, and ${\rm Reg}(\cdot)$ stands for a regularization term.   

To learn the augmented self-representation matrix based on the recovered latent representation, we combine Eq. (\ref{eq07}) with Eq. (\ref{eq08}), and arrive at
\begin{equation}
\label{eq09}
    \min_{\mathbf{P},\mathbf{H}_a,\mathbf{Z}_a} J_1\left(\mathbf{X}_a,\mathbf{PH}_a \right)+\lambda_1J_2\left(\mathbf{H}_a,\mathbf{H}_a\mathbf{Z}_a \right)+\lambda_2{\rm{Reg}}\left(\mathbf{Z}_a\right),
\end{equation}
where $\lambda_1$ and $\lambda_2$ are constants used for balancing the terms. For the purpose of avoiding redundant calculations of consistent information between different views, we consider enforcing a sparse regularization for the non-diagonal blocks of $\mathbf Z_a$. Simultaneously, to ensure the algorithm's robustness against outliers, the final objective function is presented as follows
\begin{equation}
\label{eq10}
\begin{aligned}
&\min_{\mathbf{P},\mathbf{H}_a,\mathbf{Z}_a,\mathbf{E}_1,\mathbf{E}_2} \Vert \mathbf{E}_1 \Vert_{2,1} + \lambda_1\Vert \mathbf{E}_2 \Vert_{2,1} + \lambda_2\Vert \mathbf{Z}_a-{\rm{diag}}(\mathbf{Z}_a) \Vert_1\\
    &s.t. \ \mathbf{X}_a =\mathbf{PH}_a+\mathbf{E}_1,\mathbf{H}_a=\mathbf{H}_a\mathbf{Z}_a+\mathbf{E}_2,\mathbf{P}\mathbf{P}^T=\mathbf{I},
\end{aligned}
\end{equation}
where $\mathbf E_1$, $\mathbf E_2$ are the error matrices, $\lVert\cdot\rVert_{2,1}$ and $\lVert\cdot\rVert_{1}$ perform the $l_{2,1}$-norm and $l_1$-norm operations, respectively. Alternatively, by stacking $\mathbf E_1$ and $\mathbf E_2$ into an integrated error matrix, the objective function in Eq. (\ref{eq10}) can be written into a compact form
\begin{equation}
\label{eq11}
\begin{aligned}
&\min_{\mathbf{P},\mathbf{H}_a,\mathbf{Z}_a,\mathbf{E}_1,\mathbf{E}_2} \Vert \mathbf{E} \Vert_{2,1} + \lambda\Vert \mathbf{Z}_a-{\rm{diag}}(\mathbf{Z}_a) \Vert_1\\
    &s.t. \ \mathbf{X}_a =\mathbf{PH}_a+\mathbf{E}_1,\mathbf{H}_a=\mathbf{H}_a\mathbf{Z}_a+\mathbf{E}_2,\\
&\;\;\;\;\;\;\mathbf{E}=\left[\mathbf{E}_1;\mathbf{E}_2\right],\mathbf{P}\mathbf{P}^T=\mathbf{I},
\end{aligned}
\end{equation}
where $\lambda$ is a positive balance parameter.

\textbf{\emph{Remark 1}}: It is seen from Eq. (\ref{eq03}) that the LMSC method directly concatenates the data matrices from multiple views into a single matrix along the direction of dimensionality. In this approach, only the unique information from each view $\mathbf{X}^{(l)}$, often referred to as complementary information, is utilized, while shared information, also known as consistent information, is neglected. In contrast, our method arranges the data matrices from different views within the block-diagonal segments, with the non-block-diagonal regions stacked by the correlation matrix $\mathbf{X}^{(p,q)}$ to capture inter-view similarities. Our proposed formulation can simultaneously exploit both intra-view and inter-view information for a more comprehensive latent representation recovery. Moreover, our model benefits from sparse regularization on the non-diagonal blocks, allowing it to avoid redundant inter-view information. As a result, the model’s generalization ability is improved.

\textbf{\emph{Remark 2}}: We can relate our data construction strategy with attention mechanism. From (\ref{eq04}), the non-block-diagonal data construction can be elegantly interpreted from an ``attention" perspective. Although there exits some differences -- our method, for instance, does not require dynamic updates of the data matrix -- the concept is analogous. For example, $\{\mathbf{X}^{(1,2)}, \mathbf{X}^{(1,3)},\cdots, \mathbf{X}^{(1,v)}\}$ can be interpreted as the $1$st-view garnering attention from the $\{2$rd-view ,$\cdots$,$v$th-view$\}$.

\subsection{Optimization}

To solve the optimization problem in Eq. (\ref{eq11}), we employ ADMM which is a widely used method known as the Augmented Lagrange Multiplier (ALM) with Alternating Direction Minimizing (ADM) strategy \cite{lin2011linearized}, which is suitable for addressing multivariate constrained optimization problems. The basic idea of ADM is to solve each variable induced subproblem by maintaining the other variables fixed. Before further proceeding, we need to introduce an auxiliary variable to make our cost function separable. In particular, we use $\mathbf J$ to replace $\mathbf Z_a$ in the sparse regularization term, yielding
\begin{equation}
\label{eq012}
\begin{aligned}
    &\min_{\mathbf{P},\mathbf{H}_a,\mathbf{Z}_a,\mathbf{J},\mathbf{E}_1,\mathbf{E}_2} \Vert \mathbf{E} \Vert_{2,1} + \lambda\Vert \mathbf{J}-{\rm{diag}}(\mathbf{J}) \Vert_1\\
    &s.t. \ \mathbf{X} =\mathbf{PH}_a+\mathbf{E}_1,\mathbf{H}_a=\mathbf{H}_a\mathbf{Z}_a+\mathbf{E}_2,\\
    &\;\;\;\;\;\;\mathbf{E}=\left[\mathbf{E}_1;\mathbf{E}_2\right],\mathbf{P}\mathbf{P}^T=\mathbf{I},\mathbf{J}=\mathbf{Z}_a.
\end{aligned}
\end{equation}

With the ALM method, we have from Eq. (\ref{eq012})
\begin{equation}
\label{eq013}
\begin{aligned}
&\mathcal{L}\left(\mathbf{P},\mathbf{H}_a,\mathbf{Z}_a,\mathbf{E}_1,\mathbf{E}_2,\mathbf{J} \right)\\
    &=\Vert \mathbf{E} \Vert_{2,1}+\lambda\Vert \mathbf{J}-{\rm{diag}}(\mathbf{J}) \Vert_1+\Phi\left(\mathbf{Y}_1,\mathbf{X}_a-\mathbf{PH}_a-\mathbf{E}_1\right)\\
    &\qquad +\Phi\left(\mathbf{Y}_2,\mathbf{H}_a-\mathbf{H}_a\mathbf{Z}_a-\mathbf{E}_2\right)+\Phi\left(\mathbf{Y}_3,\mathbf{J}-\mathbf{Z}_a\right) \\
    &\quad s.t. \ \mathbf{PP}^T=\mathbf{I},
\end{aligned}
\end{equation}
where $\Phi\left(\mathbf{\Theta},\mathbf{\Psi}\right)=\frac{\mu}{2}\Vert \mathbf{\Psi} \Vert_F^2+{\rm{Tr}}\left(\mathbf{\Theta}^T\mathbf{\Psi} \right)$ with $\mu$ being a positive value, $\mathbf Y_1$, $\mathbf Y_2$ and $\mathbf Y_3$ are Lagrange multipliers. To solve Eq. (\ref{eq013}), we decompose our problem into the subsequent subproblems. 

\textbf{1) P-subproblem}

To update $\mathbf{P}$, we treat $\mathbf P$ as a variable while maintaining the other variables fixed, leading to
\begin{equation}
\label{eq014}
\begin{aligned}
    &\mathbf{P}^* = \arg \min \Phi\left(\mathbf{Y}_1,\mathbf{X}_a-\mathbf{PH}_a-\mathbf{E}_1\right) \\
    &s.t. \ \mathbf{PP}^T=\mathbf{I}.
\end{aligned}
\end{equation}
The objective optimization in Eq. (\ref{eq014}) can be equivalently transformed into
\begin{equation}
\label{eq015}
\begin{aligned}
\mathbf P^* &= \arg \min \frac{\mu}{2}\lVert\mathbf X_a-\mathbf P\mathbf {H}_a-\mathbf{E}_1+\mathbf{Y}_1\rVert^2_F\\
&= \arg \min \frac{\mu}{2}\lVert(\mathbf{X}_a+\mathbf{Y}_1/\mu-\mathbf{E}_1)-\mathbf{PH}_a\rVert^2_F\\
&= \arg\min \frac{\mu}{2}\lVert(\mathbf{X}_a\mathbf{Y}_1/\mu-\mathbf{E}_1)^T-\mathbf{H}^T_a\mathbf{P}^T\rVert^2_F.
\end{aligned}
\end{equation}

\noindent To further proceed with Eq. (\ref{eq015}), it is necessary to present the following theorem.

\textbf{\emph{Theorem 1}} \cite{huang2013spectral}: For the objective function  ${\rm min}_{\mathbf \Gamma}\lVert\mathbf\Upsilon-\mathbf\Xi\mathbf\Gamma\rVert^2_F$ $s.t.$ $\mathbf\Gamma^T\mathbf\Gamma=\mathbf\Gamma\mathbf\Gamma^T=\mathbf I$, the optimal solution is $\mathbf\Gamma=\mathbf U\mathbf V^T$, where $\mathbf U$ and $\mathbf V$ are left and right singular values of SVD decomposition of $\mathbf\Xi^T\mathbf\Upsilon$. 

According to the above theorem, it is not difficult to derive that the solution to Eq. (\ref{eq015}) is $\mathbf{P}^T=\mathbf{UV}^T$, where $\mathbf{U}$ and $\mathbf{V}$ is the left and right singular values of SVD of $\mathbf{H}_a(\mathbf{Y}_1/\mu+\mathbf{X}_a-\mathbf{E}_1)^T$.

\textbf{2)} $\mathbf H_a$\textbf{-subproblem}

To update $\mathbf{H}_a$, we treat $\mathbf{H}_a$ as a variable while keeping the other variables fixed, resulting in
\begin{equation}
\label{eq016}
\begin{aligned}
    &\mathbf{H}_a^* = \arg \min \Phi\left(\mathbf{Y}_1,\mathbf{X}_a-\mathbf{PH}_a-\mathbf{E}_1\right)\\
    &\qquad\qquad\qquad+\Phi\left(\mathbf{Y}_2,\mathbf{H}_a-\mathbf{H}_a\mathbf{Z}_a-\mathbf{E}_2\right).
\end{aligned}
\end{equation}
Taking the derivative with respect to $\mathbf{H}_a$ and setting the resulted equation to zero, we arrive at
\begin{equation}
\label{eq017}
\begin{aligned}
    &\mathbf{A}\mathbf{H}_a+\mathbf{H}_a\mathbf{B}=\mathbf{C}\\
    {\rm{with}} \ &\mathbf{A}=\mu\mathbf{P}^T\mathbf{P},\\
    &\mathbf{B}=\mu\left(\mathbf{Z}_a\mathbf{Z}_a^T-\mathbf{Z}_a-\mathbf{Z}_a^T+\mathbf{I} \right),\\
    &\mathbf{C}=\mathbf{P}^T\mathbf{Y}_1+\mathbf{Y}_2\left(\mathbf{Z}_a^T-\mathbf{I}\right)\\
    & \quad +\mu\left(\mathbf{P}^T\mathbf{X}_a+\mathbf{E}_2^T-\mathbf{P}^T\mathbf{E}_1-\mathbf{E}_2\mathbf{Z}_a^T \right).
\end{aligned}
\end{equation}
The equation presented above is a Sylvester equation, and it can be effectively solved using the ``lyap" function in MATLAB.

\textbf{3)} $\mathbf Z_a$\textbf{-subproblem}

Upon fixing the other variables, we update $\mathbf Z_a$ by solving the following objective optimization problem 
\begin{equation}
\label{eq018}
\begin{aligned}
    \mathbf{Z}_a^* = \arg \min \Phi\left(\mathbf{Y}_3,\mathbf{J}-\mathbf{Z}_a\right)+\Phi\left(\mathbf{Y}_2,\mathbf{H}_a-\mathbf{H}_a\mathbf{Z}_a-\mathbf{E}_2\right).
\end{aligned}
\end{equation}
By taking the derivative with respect to $\mathbf{Z}_a$ and setting the resulted equation to zero, we obtain
\begin{equation}
\begin{aligned}
\label{eq019}
    \mathbf{Z}_a^*=&\left(\mathbf{H}_a^T\mathbf{H}_a+\mathbf{I} \right)^{-1}\\
    &\qquad\left[\left(\mathbf{J}+\mathbf{H}_a^T\mathbf{H}_a-\mathbf{H}_a^T\mathbf{E}_2\right)+\left(\mathbf{Y}_3+\mathbf{H}_a^T\mathbf{Y}_2 \right)/\mu \right].
\end{aligned}
\end{equation}

\textbf{4) E-subproblem}

Under the condition of fixing other variables, the error matrix $\mathbf{E}$ is updated by solving the following problem
\begin{equation}
\label{eq020}
\begin{aligned}
    \mathbf{E}^*&=\arg \min \Vert \mathbf{E} \Vert_{2,1}+\Phi\left(\mathbf{Y}_1,\mathbf{X}_a-\mathbf{PH}_a-\mathbf{E}_1\right)\\
    &\qquad\qquad\qquad\qquad\qquad+\Phi\left(\mathbf{Y}_2,\mathbf{H}_a-\mathbf{H}_a\mathbf{Z}_a-\mathbf{E}_2\right) \\
    &=\frac{1}{\mu}\Vert \mathbf{E} \Vert_{2,1}+\frac{1}{2}\Vert \mathbf{E-G} \Vert_F^2,
\end{aligned}
\end{equation}
where $\mathbf{G}=\left[\mathbf{X}_a-\mathbf{PH}_a+\mathbf{Y}_1/\mu;\mathbf{H}_a-\mathbf{H}_a\mathbf{Z}_a+\mathbf{Y}_2/\mu \right]$. This subproblem can be effectively solved by using Lemma 4.1 in \cite{liu2012robust}.

\textbf{5) J-subproblem}

Following a similar approach, we update $\mathbf{J}$ by solving the following optimization problem 
\begin{equation}
\label{eq021}
\begin{aligned}
    \mathbf{J}^*&=\arg \min \lambda\Vert \mathbf{J}-{\rm{diag}}(\mathbf{J}) \Vert_1+\Phi\left(\mathbf{Y}_3,\mathbf{J}-\mathbf{Z}_a\right) \\
    &=\arg \min \frac{\lambda}{\mu}\Vert \mathbf{J}-{\rm{diag}}(\mathbf{J}) \Vert_1+\frac{1}{2}\Vert \mathbf{J}-\left(\mathbf{Z}_a-\mathbf{Y}_3/\mu \right) \Vert_F^2.
\end{aligned}
\end{equation}
By using the singular value thresholding operator \cite{cai2010singular}, the update on $\mathbf J$ has a closed-form solution given by
\begin{equation}
\label{eq022}
\begin{aligned}
    &\mathbf{J}^*=\mathbf{J}_1+\mathbf{J}_2 \\
    {\rm{with}} \ &\mathbf{J}_1={\rm{blkdiag}}(\mathbf{J})  \\
    &\mathbf{J}_2=\mathcal{T}_{\frac{\lambda}{\mu}}\left(\mathbf{J}-\mathbf{J}_1 \right)\\
    &\mathbf{J}=\mathbf{Z}_a-\mathbf{Y}_3/\mu,
\end{aligned}
\end{equation}
where {\rm blkdiag}$(\cdot)$ takes the diagonal blocks of a matrix, and $\mathcal{T}(\cdot)$ is the shrinage-thresholding operator acting on each element of a given matrix for solving sparse problem, defined by
\begin{equation}
\label{eq023}
\begin{aligned}
    \mathcal{T}_{\eta}(\gamma)=\left(|\gamma|-\eta \right)_{+}{\rm{sgn}}(\gamma).
\end{aligned}
\end{equation}

\textbf{6) Updating multipliers}

By applying the gradient ascent operation, we update the Lagrange multipliers as 
\begin{equation}
\label{eq024}
    \left\{
    \begin{aligned}
        &\mathbf{Y}_1=\mathbf{Y}_1+\mu\left(\mathbf{X}_a-\mathbf{PH}_a-\mathbf{E}_1\right) \\
        &\mathbf{Y}_2=\mathbf{Y}_2+\mu\left(\mathbf{H}_a-\mathbf{H}_a\mathbf{Z}_a-\mathbf{E}_2\right) \\
        &\mathbf{Y}_3=\mathbf{Y}_3+\mu\left(\mathbf{J}-\mathbf{Z}_a\right) \\
    \end{aligned}
    \right.
\end{equation}
where $\mu=\min(\rho\mu,\mu_{max})$ with $\rho$ being a constant. The procedures of implementing ELMSC are summarized in Algorithm \ref{alg: learning Z_a}. 

\begin{algorithm}[tb]
\caption{Procedures of implementing ELMSC}
\label{alg: learning Z_a}
\textbf{Input: }Multi-view data matrices $\{\mathbf{X}^{(1)},\cdots,\mathbf{X}^{(v)}\}$, \\
parameter $\lambda$ and the dimension $k$ of $\mathbf H_a$.\\
\textbf{Initialize: }Initialize $\mathbf{H}_a$ using the standard Gaussian distribution; $\mathbf{P=0}$, $\mathbf{E}_1=\mathbf{0}$, $\mathbf{E}_2=\mathbf{0}$, $\mathbf{Z}_a=\mathbf{J=0}$, $\mathbf{Y}_1=\mathbf{0}$, $\mathbf{Y}_2=\mathbf{0}$, $\mathbf{Y}_3=\mathbf{0}$, $\mu=10^{-4}$, $\mu_{max}=10^6$, $\rho=1.2$, $tol=10^{-3}$, $t=0$, $T=100$.
\begin{algorithmic}[1] 
\STATE Calculate $\mathbf{X}_a$ according to Eqs. (\ref{eq04})-(\ref{eq06}).
\WHILE{not converged and $t<T$}
\STATE Update variable $\mathbf{P}$ by solving Eq. (\ref{eq015}).
\STATE Update variable $\mathbf{H}_a$ by Eq. (\ref{eq017}).
\STATE Update variable $\mathbf{Z}_a$ by Eq. (\ref{eq019}).
\STATE Update variable $\mathbf{E}$ by solving Eq. (\ref{eq020}).
\STATE Update variable $\mathbf{J}$ by Eq. (\ref{eq022}).
\STATE Update multipiers $\mathbf{Y}_1$, $\mathbf{Y}_2$, $\mathbf{Y}_3$ according to Eq. (\ref{eq024}).
\STATE Update $\mu$ by $\mu=\min(\rho\mu,\mu_{max})$.
\IF {$\Vert\mathbf{X}_a-\mathbf{PH}_a-\mathbf{E}_1\Vert_{\infty}<tol$ and $\Vert\mathbf{H}_a-\mathbf{H}_a\mathbf{Z}_a-\mathbf{E}_2\Vert_{\infty}<tol$ and $\Vert\mathbf{J}-\mathbf{Z}_a\Vert_{\infty}<tol$}
\STATE Converged.
\ENDIF
\STATE $t\gets t+1$.
\ENDWHILE
\end{algorithmic}
\textbf{Output: }$\mathbf{Z}_a$, $\mathbf{H}_a$, $\mathbf{P}$, $\mathbf{E}$
\end{algorithm}

\textbf{\emph{Remark 3}}: It should be noted that in contrast to the conventional self-representation matrix of dimensions $n\times n$, the augmented self-representation matrix denoted as $\mathbf Z_a$ in our proposed ELMSC, has dimensions of $vn\times vn$. As a result, it is not feasible to directly utilize $\mathbf Z_a$ to calculate the affinity matrix. Given that $\mathbf Z_a$ comprises $v^2$ individual block matrices, we arrive at the final self-representation matrix by accumulating all these block matrices. This procedure can be expressed as:
\begin{equation}
\label{eq025}
\hat{\mathbf{Z}}=\sum^v_i\sum^v_j\mathbf{Z}_a(i,j),
\end{equation}
where $\mathbf{Z}_a(i,j)$ denotes the $(i,j)$-th block matrix. 

It is justifiable to use Eq. (\ref{eq025}) for calculating the final self-representation matrix, the rationale for which is provided below. Recall that we impose sparse regularization on non-diagonal blocks. Besides its primary purpose in circumventing redundant computations, this regularization serves an additional purpose: it encourages sparsity, indicating limited contributions from non-diagonal blocks. This implies that our design assigns higher importance to diagonal blocks and lesser to non-diagonal blocks, actually reflecting the contribution weights of each block.  

Based on the resulted $\hat{\mathbf{Z}}$, one can perform spectral clustering to obtain the final clustering results.

\subsection{Computational Complexity}

Our method mainly consists of two stages: 1) construction of $\mathbf{X}_a$ ; 2) updates of various subproblems. In the first stage, we require to calculate Eq. (\ref{eq05}) and Eq. (\ref{eq06}), which takes a complexity of $O(v^2n^2d)$. In the second stage, we need to update subproblems 1-6. Specifically, the update for $\mathbf P$ involves the calculation of $\mathbf{H}_a(\mathbf{Y}_1/\mu + \mathbf{X}_a-\mathbf{E}_1)$, the execution of SVD on the resulting matrix, and the computation of $\mathbf{U}\mathbf{V}^T$, which in total consumes $O(kd^2+kvnd)$. The update for $\mathbf{H}_a$ incorporates the updates of $\mathbf{A}$, $\mathbf{B}$ and $\mathbf{C}$, as well as solving a Sylvester equation. The total complexity for this is $O((vn)^3+k^3+k^2d+kdvn)$. The update for $\mathbf{Z}_a$ has a complexity of $O((vn)^3+(vn)^2k)$. Finally, the updates for $\mathbf{E}$ and $\mathbf{J}$ have complexities of $O(dkvn+k(vn)^2)$ and $O((vn)^2)$, respectively. 

In summary, the overall complexity is $O(v^2n^2d+v^3n^3+v^2n^2k+kdvn+v^2n^2+d^2k+k^3)$. Generally, the conditions $k\ll d$, $v\ll n$ and $d\ll n$ hold true. Therefore, the approximate complexity can be regarded as $O(v^3n^3)$, which preserves the complexity order of $O(n^3)$ that is comparable to many prevalent tensor-based multi-view methods \cite{tang2021constrained, xia2021multiview}. In future work, we may explore scalable methods to reduce computational complexity \cite{kang2021structured, pan2023high}. 

\section{Theoretical Study}

\subsection{Connection between LMSC and Our Method}

In contrast to LMSC, our method innovatively introduces an enhanced data reconstruction strategy, which particularly emphasizes the correlation matrix $\mathbf{X}^{(p,q)}$ between two different views. Despite the existence of a clear distinction between LMSC and our method, we can establish the relationship between them by performing the following discussion.

If we replace the cosine similarity matrix $\mathbf{S}^{(p,q)}$ with an identity matrix, the augmented data matrix $\mathbf{X}_a$ is transformed into
\begin{equation}
    \label{eq026}
    \mathbf X_a = \left[
    \begin{array}{c c c}
        \mathbf X^{(1)}, & \cdots, & \mathbf X^{(1)} \\
        \mathbf X^{(2)}, & \cdots, & \mathbf X^{(2)} \\
        \vdots & \vdots & \vdots\\
        \mathbf X^{(v)}, & \cdots, & \mathbf X^{(v)}\\
    \end{array}
    \right].
\end{equation}
Given that $\mathbf{X}_a = \mathbf{P}\mathbf{H}_a + \mathbf{E}_1$, we have the following equation by ignoring the error matrix $\mathbf{E}_1$
\begin{equation}
    \label{eq027}
    \left[
    \begin{array}{c c c c}
        \mathbf X^{(1)}, & \cdots, & \mathbf X^{(1)} \\
        \mathbf X^{(2)}, & \cdots, & \mathbf X^{(2)} \\
        \vdots & \vdots & \vdots\\
        \mathbf X^{(v)} & \cdots, & \mathbf X^{(v)}\\
    \end{array}
    \right] = 
    \left[ \begin{array}{c}
         \mathbf{P}^{(1)} \\
         \vdots\\
         \mathbf{P}^{(v)} \\
    \end{array}
    \right]
    \left[ \begin{array}{c c c}
         \mathbf{H}^{(1)}, & \cdots, & \mathbf{H}^{(v)} \\
    \end{array}
    \right].
\end{equation}
which consequently results in $\mathbf{H}^{(1)} = \cdots = \mathbf{H}^{(v)}$.

It is noted that in our objective function, we enforce a sparse regularization on the non-diagonal blocks of $\mathbf{Z}_a$, denoted as $\Vert \mathbf{Z}_a-{\rm{diag}}(\mathbf{Z}_a) \Vert_1$. Assuming this sparse regularization is strictly satisfied, that is,
\begin{equation}
	\label{eq028}
	\mathbf{Z}_a = 
	\left[
	\begin{array}{c c c}
		\mathbf{Z}^{(1)}, & \cdots, &\mathbf{0}\\
		\mathbf{0}, & \cdots, &\mathbf{0}\\
		\vdots & \vdots & \vdots\\	
		\mathbf{0}, & \cdots, &\mathbf{Z}^{(v)}					
	\end{array}	
	\right],
\end{equation}
and since $\mathbf{H}_a = \mathbf{H}_a\mathbf{Z}_a + \mathbf{E}_2$, we have the following result by ignoring the error matrix $\mathbf{E}_2$
\begin{equation}
\begin{aligned}
	\label{eq029}
	 &\left[ \begin{array}{c c c}
         \mathbf{H}^{(1)}, & \cdots, & \mathbf{H}^{(v)} \\
    \end{array}
    \right]\\
    &\qquad\qquad = 
	 \left[ \begin{array}{c c c}
         \mathbf{H}^{(1)}, & \cdots, & \mathbf{H}^{(v)} \\
    \end{array}
    \right]
    \left[
	\begin{array}{c c c}
		\mathbf{Z}^{(1)}, & \cdots, &\mathbf{0}\\
		\mathbf{0}, & \cdots, &\mathbf{0}\\
		\vdots, & \vdots & \vdots\\	
		\mathbf{0}, & \cdots, &\mathbf{Z}^{(v)}					
	\end{array}	
	\right].
\end{aligned}
\end{equation}
From Eq. (\ref{eq029}), it is obvious that
\begin{equation}
	\label{eq030}
	  \left\{
    \begin{aligned}
        &\mathbf{H}^{(1)}=\mathbf{H}^{(1)} \mathbf{Z}^{(1)} \\
        &\mathbf{H}^{(2)}=\mathbf{H}^{(2)} \mathbf{Z}^{(2)} \\
		& \cdots\\        
        &\mathbf{H}^{(v)}=\mathbf{H}^{(v)} \mathbf{Z}^{(v)}. \\
    \end{aligned}
    \right.
\end{equation}

Consequently, we obtain $\mathbf{Z}^{(1)}=\mathbf{Z}^{(2)} = \cdots = \mathbf{Z}^{(v)}$, which is consistent with the latent representation in LMSC. From the above analysis, we demonstrate that our method can be regarded as an augmented variant of LMSC. Furthermore, under specific conditions, it simplifies to LMSC.

\subsection{Convergence Analysis}

Let us define $\mathbf{Y} = [\mathbf{Y}_{1}; \mathbf{Y}_{2}; \mathbf{Y}_{3}]$ and $\mathbf{Y}_{12} = [\mathbf{Y}_1; \mathbf{Y}_2]$. Then, it is natural to have $\mathbf{Y}_t = [\mathbf{Y}_{1,t}; \mathbf{Y}_{2,t}; \mathbf{Y}_{3,t}]$ and $\mathbf{Y}_{12,t} = [\mathbf{Y}_{1,t}; \mathbf{Y}_{2,t}]$. To prove the convergence of ELMSC, we need to complete the proof of the following theorem. 

\noindent \textbf{\emph{Theorem 1}.} Suppose $\{\boldsymbol{\zeta}_t = (\mathbf{P}_t, \mathbf{H}_{a,t}, \mathbf{Z}_{a,t}, \mathbf{E}_{t}, \mathbf{J}_{t}; \mathbf{Y}_t) \}^\infty_{t=1}$ be the sequence generated from {\bf{Algorithm 1}}, we state that ELMSC is convergent since the sequence $\{\boldsymbol{\zeta}_t\}$ satisfies the following properties: 

1) The sequence $\{\boldsymbol{\zeta}_t\}^\infty_{t=1}$ is bounded.

2) The sequence $\{\boldsymbol{\zeta}_t\}^\infty_{t=1}$ has at least one accumulation point, denoted as $\boldsymbol{\zeta}_*$. The accumulation point $\boldsymbol{\zeta}_*$ is a stationary point as it satisfies the first-order optimality Karush-Kuhn-Tucker (KKT) conditions \cite{guo2022logarithmic}:
\begin{equation}
\label{eq31}
    \left\{
    \begin{aligned}
        &\mathbf{X}_a - \mathbf{P}_*\mathbf{H}_{a,*} - \mathbf{E}_{1,*} = \mathbf{0}, \\
        &\mathbf{H}_{a,*} - \mathbf{H}_{a,*}\mathbf{Z}_{a,*} - \mathbf{E}_{2,*} = \mathbf{0}, \\
        &\mathbf{J}_* - \mathbf{Z}_{a,*} = \mathbf{0}, \\
        &\mathbf{Y}_{12,*}\in\partial_{\mathbf{E}}\lVert\mathbf{E}_{*}\rVert_{2,1}, \\
        &\mathbf{Y}_{3,*} \in \partial_{\mathbf{J}}\lVert\mathbf{J}_{*}-{\rm{diag}}{(\mathbf{J}_*)}\rVert_1. 
    \end{aligned}
    \right.
\end{equation}

\noindent \textbf{\emph{Proof}}: 

1) Proof of the boundness of $\{\mathbf{Y}_{1,t}\}$, $\{\mathbf{Y}_{2,t}\}$ and $\{\mathbf{Y}_{3,t}\}$.

The \textbf{E-subproblem} is equivalent to optimize:
\begin{equation}
\label{eq32}
	\arg \min \Vert \mathbf{E} \Vert_{2,1}+\frac{\mu}{2}\Vert \mathbf{E-G} \Vert_F^2.
\end{equation}

\noindent Suppose at iteration $t$, the above optimization problem can be divided into two problems, as follows:
\begin{equation}
\begin{aligned}
\label{eq33}
	\mathbf{E}_{1, t+1} = &\arg \min \Vert \mathbf{E}_1 \Vert_{2,1}\\
	&+\frac{\mu_t}{2}\Vert \mathbf{E}_{1}-(\mathbf{X}_a-\mathbf{P}_{t+1}\mathbf{H}_{a,t+1}+\mathbf{Y}_{1,t}/\mu_t) \Vert_F^2,
\end{aligned}
\end{equation}
and
\begin{equation}
\begin{aligned}
\label{eq34}
	\mathbf{E}_{2, t+1} = &\arg \min \Vert \mathbf{E}_2 \Vert_{2,1}\\
	&+\frac{\mu_t}{2}\Vert \mathbf{E}_2-(\mathbf{H}_{a,t+1}-\mathbf{H}_{a,t+1}\mathbf{Z}_{a,t+1}+\mathbf{Y}_{2,t}/\mu_t) \Vert_F^2.
\end{aligned}
\end{equation}

The solution of \textbf{E-subproblem} follows from Lemma 4.1 in \cite{liu2012robust}. It is easy to know that the optimal $\mathbf{E}_{1,t+1}$ satisfies the first-order optimality condition, i.e.,
\begin{equation}
\label{eq35}
	\mathbf{0} \in \partial\Vert\mathbf{E}_{1, t+1}\Vert_{2,1} + \mu_t[\mathbf{E}_{1,t+1}-(\mathbf{X}_a-\mathbf{P}_{t+1}\mathbf{H}_{a,t+1}+\mathbf{Y}_{1,t}/\mu_t)].
\end{equation}

\noindent By employing the updating rule of $\mathbf{Y}_{1,t}$, Eq.({\ref{eq35}}) can be further expressed as
\begin{equation}
\label{eq36}
	\mathbf{0} \in \partial\Vert\mathbf{E}_{1, t+1}\Vert_{2,1} - \mathbf{Y}_{1,t+1}.
\end{equation}

\noindent It is straightforward to see that $\partial\Vert\mathbf{E}_{1, t+1}\Vert_{2,1}$ is bounded since 
\begin{equation}
\label{eq37}
    \partial_{\mathbf{E}_{1,t+1}}(\lVert\mathbf{E}_{1}\rVert_{2,1}) = \left\{
    \begin{aligned}
        &0, & if \;\lVert\mathbf{E}_{1,t+1}(:,i)\rVert_2 = 0\\
        &\frac{\mathbf{E}_{1,t+1}(:,i)}{\lVert\mathbf{E}_{1,t+1}(:,i)\rVert_2}, & otherwise. 
    \end{aligned}
    \right.
\end{equation}

\noindent Therefore, $\mathbf{Y}_{1,t+1} \in \partial\Vert\mathbf{E}_{1, t+1}\Vert_{2,1}$, then $\{\mathbf{Y}_{1,t}\}$ is bounded. Following similar procedures, we can prove that $\{\mathbf{Y}_{2,t}\}$ and $\{\mathbf{Y}_{3,t}\}$ are bounded. Note that when proving the boundness of $\{\mathbf{Y}_{3,t}\}$, some useful properties can be referenced from \cite{wang2019global}. 

2) Proof of the boundness of $\{\mathbf{P}_t\}$, $\{\mathbf{H}_{a,t}\}$, $\{\mathbf{Z}_{a,t}\}$, $\{\mathbf{E}_{t}\}$, $\{\mathbf{J}_t\}$.

The detailed proof the boundness of $\{\mathbf{P}_t\}$, $\{\mathbf{H}_{a,t}\}$, $\{\mathbf{Z}_{a,t}\}$, $\{\mathbf{E}_{t}\}$, $\{\mathbf{J}_t\}$ can be found in the supplementary material. Based on the proofs 1) and 2), we can conclude that $\{\boldsymbol{\zeta}_t\}^\infty_{t=1}$ is bounded.

3) Proof of the existing of stationary point.

According to Bolzano-Weierstrass theorem \cite{bartle2000introduction}, a bounded sequence has at least one convergent sequence which converges to an accumulation point. Suppose that $\{\boldsymbol{\zeta}_t\}$ exists an accumulation point, denoted by $\boldsymbol{\zeta}_* = \{\mathbf{P}_*, \mathbf{H}_{a,*}, \mathbf{Z}_{a,*}, \mathbf{E}_*, \mathbf{J}_*, \mathbf{Y}_*\}$. Thus, we have
\begin{equation}
\label{eq38}
\lim_{t\rightarrow\infty}(\mathbf{P}_t, \mathbf{H}_{a,t}, \mathbf{Z}_{a,t}, \mathbf{E}_t, \mathbf{J}_t, \mathbf{Y}_t) = (\mathbf{P}_*, \mathbf{H}_{a,*}, \mathbf{Z}_{a,*}, \mathbf{E}_*, \mathbf{J}_*, \mathbf{Y}_*).
\end{equation}

From the updating rules of Lagrange multipliers, we arrive at 
\begin{equation}
\label{eq39}
    \left\{
    \begin{aligned}
        &(\mathbf{Y}_{1,t+1}-\mathbf{Y}_{1,t})/\mu_t = \mathbf{X}_a - \mathbf{P}_{t+1}\mathbf{H}_{a,t+1} - \mathbf{E}_{1,t+1}\\
        &(\mathbf{Y}_{2,t+1}-\mathbf{Y}_{2,t})/\mu_t = \mathbf{H}_{a,t+1}-\mathbf{H}_{a,t+1}\mathbf{Z}_{a,t+1}-\mathbf{E}_{2,t+1}\\
        &(\mathbf{Y}_{3,t+1}-\mathbf{Y}_{3,t})/\mu_t = \mathbf{J}_{t+1}-\mathbf{Z}_{a,t+1}   
    \end{aligned}
    \right.
\end{equation}

\noindent With the boundness of $\{\mathbf{Y}_{1,t}\}$, $\{\mathbf{Y}_{2,t}\}$ and $\{\mathbf{Y}_{3,t}\}$ and $\lim_{t\rightarrow\infty} = \infty$, we yield 
\begin{equation}
\label{eq40}
    \left\{
    \begin{aligned}
        &\lim_{t\rightarrow\infty} \mathbf{X}_a - \mathbf{P}_{t+1}\mathbf{H}_{a,t+1} - \mathbf{E}_{1,t+1} = \mathbf{0}\\
        &\lim_{t\rightarrow\infty} \mathbf{H}_{a,t+1}-\mathbf{H}_{a,t+1}\mathbf{Z}_{a,t+1}-\mathbf{E}_{2,t+1} = \mathbf{0}\\
        &\lim_{t\rightarrow\infty} \mathbf{J}_{t+1}-\mathbf{Z}_{a,t+1} = \mathbf{0}   
    \end{aligned}
    \right.
\end{equation}
thus,
\begin{equation}
\label{eq41}
    \left\{
    \begin{aligned}
        & \mathbf{X}_a - \mathbf{P}_{*}\mathbf{H}_{a,*} - \mathbf{E}_{1,*} = \mathbf{0}\\
        &\mathbf{H}_{a,*}-\mathbf{H}_{a,*}\mathbf{Z}_{a,*}-\mathbf{E}_{2,*} = \mathbf{0}\\
        &\mathbf{J}_{*}-\mathbf{Z}_{a,*} = \mathbf{0}   
    \end{aligned}
    \right.
\end{equation}

Consider that the optimal $\mathbf{E}_{t+1}$, $\mathbf{J}_{t+1}$ satisfy the first-order optimality conditions, i.e.,
\begin{equation}
\label{eq42}
	\mathbf{0} \in \partial\Vert\mathbf{E}_{t+1}\Vert_{2,1} - \mathbf{Y}_{12,t+1} \Rightarrow \mathbf{Y}_{12,*}\in\partial\Vert\mathbf{E}_{*}\Vert_{2,1}\;\;\; for\;\; {t\rightarrow\infty}
\end{equation}

\begin{equation}
\begin{aligned}
\label{eq43}
	\mathbf{0} \in \lambda \partial\Vert\mathbf{J}_{t+1}-&{\rm{diag}}(\mathbf{J}_{t+1})\Vert_{1} - \mathbf{Y}_{3,t+1}\\
	&\Rightarrow \mathbf{Y}_{3,*}\in\partial\Vert\mathbf{J}_{*}-{\rm{diag}}(\mathbf{J}_{*})\Vert_{1}\;\;\; for\;\; {t\rightarrow\infty}
\end{aligned}
\end{equation}

\noindent With the above analysis, we conclude that the accumulation point $\boldsymbol{\zeta}_*$ satisfies the first-order optimality KKT conditions, therefore it is a stationary point.

\section{Experiments}

\begin{table*}[h!]
    \centering
    \tiny
    \caption{Evaluation metrics of multi-view clustering methods on ORL, MSRCV1 and HW $(\%)$. Red font indicates the highest performance, blue font indicates the second best performance, and underline font indicates the third best performance.}
    \label{tab01}
    \resizebox{\linewidth}{!}{
    \begin{tabular}{c c c c c c}
         \bottomrule
         Datasets & Method & ACC & NMI & AR & F1 \\
        \hline
        & $\rm{LRR_{BestSV}}$ & $74.70\pm1.91$ & $92.50\pm0.70$ & $68.60\pm3.15$ & $69.44\pm3.05$ \\
        & Co-reg & $81.50\pm1.67$ & $90.22\pm0.53$ & $74.15\pm1.67$ & $74.76\pm1.63$ \\
        & AWP & $79.25\pm0.00$ & $90.11\pm0.00$ & $73.31\pm0.00$ & $73.95\pm0.00$ \\
        & MLRSSC & $73.53\pm1.10$ & $85.57\pm0.71$ & $62.60\pm1.28$ & $63.50\pm1.25$ \\
        & MCGC & $73.00\pm0.00$ & $86.86\pm0.00$ & $62.09\pm0.00$ & $63.04\pm0.00$ \\
        & CSMSC & $81.30\pm2.06$ & $92.54\pm0.75$ & $76.27\pm2.24$ & $76.85\pm2.18$ \\
        & MCLES & $79.65\pm2.51$ & $90.75\pm1.35$ & $69.47\pm3.71$ & $70.25\pm3.59$ \\
        ORL & LMSC & $83.25\pm1.81$ & $92.77\pm0.46$ & $\underline{77.41\pm1.53}$ & $\underline{77.95\pm1.49}$ \\
        & TBGL & $73.50\pm0.00$ & $85.56\pm0.00$ & $31.96\pm0.00$ & $34.38\pm0.00$ \\
		& LTBPL & $81.75\pm0.00$ & $\underline{93.72\pm0.00}$ & $61.54\pm0.00$ & $62.67\pm0.00$\\
		& WTSNM & $\underline{84.25\pm0.00}$ & $91.17\pm0.00$ & $76.10\pm0.00$ & $76.66\pm0.00$\\ 
		& MCLGF & $82.75\pm0.00$ & $88.93\pm0.00$ & $72.70\pm0.00$ & $73.32\pm0.00$\\    
        & \textbf{Ours} & ${\color{red}{90.50\pm1.57}}$ & ${\color{red}{96.43\pm0.56}}$ & ${\color{red}{88.31\pm1.63}}$ & ${\color{red}{88.59\pm1.60}}$ \\
        & \textbf{Ours(random)} & ${\color{blue}{85.88\pm1.67}}$ & ${\color{blue}{95.53\pm0.44}}$ & ${\color{blue}{83.54\pm1.29}}$ & ${\color{blue}{83.93\pm1.26}}$\\
        \hline
        \hline
        & $\rm{LRR_{BestSV}}$ & $62.86\pm0.00$ & $55.23\pm0.00$ & $44.28\pm0.00$ & $53.09\pm0.00$ \\
        & Co-reg & $80.71\pm0.25$ & $72.85\pm0.63$ & $64.92\pm0.53$ & $69.88\pm0.45$ \\
        & AWP & $75.24\pm0.00$ & $72.00\pm0.00$ & $62.75\pm0.00$ & $68.66\pm0.00$ \\
        & MLRSSC & $29.76\pm0.79$ & $11.70\pm0.37$ & $4.77\pm0.33$ & $18.28\pm0.27$ \\
        & MCGC & $74.76\pm0.00$ & $67.89\pm0.00$ & $59.97\pm0.00$ & $65.88\pm0.00$ \\
        & CSMSC & $85.86\pm0.60$ & $76.54\pm0.87$ & $70.98\pm1.07$ & $75.06\pm0.91$ \\
        & MCLES & $84.57\pm6.79$ & $80.52\pm2.73$ & $72.58\pm5.46$ & $76.55\pm4.55$ \\
        MSRCV1 & LMSC & $77.76\pm2.79$ & $69.79\pm3.51$ & $62.63\pm5.02$ & $67.94\pm4.33$ \\
        & TBGL & $\underline{98.10\pm0.00}$ & $\underline{96.03\pm0.00}$ & $95.54\pm0.00$ & $96.16\pm0.00$ \\
		& LTBPL & $97.52\pm0.00$ & $97.92\pm0.00$ & ${\color{red}{97.88\pm0.00}}$ & ${\color{red}{98.03\pm0.00}}$\\
		& WTSNM & $80.00\pm0.00$ & $67.31\pm0.00$ & $59.94\pm0.00$ & $65.60\pm0.00$\\
		& MCLGF & $88.57\pm0.00$ & $80.60\pm0.00$ & $75.84\pm0.00$ & $79.24\pm0.00$\\		       
        & \textbf{Ours} & ${\color{red}{98.57\pm0.78}}$ & ${\color{red}{97.03\pm1.38}}$ & ${\color{blue}{96.68\pm1.75}}$ & ${\color{blue}{97.14\pm1.51}}$ \\
        & \textbf{Ours(random)} & ${\color{blue}{98.43\pm0.39}}$ & ${\color{blue}{96.75\pm0.70}}$ & $\underline{96.33\pm0.89}$ & $\underline{96.84\pm0.77}$\\
        \hline
        \hline
        & $\rm{LRR_{BestSV}}$ & $45.42\pm0.31$ & $40.64\pm0.17$ & $27.05\pm0.23$ & $34.76\pm0.20$ \\
        & Co-reg & $72.29\pm0.71$ & $74.86\pm1.00$ & $63.81\pm1.13$ & $67.54\pm1.03$ \\
        & AWP & $69.85\pm0.00$ & $75.76\pm0.00$ & $61.59\pm0.00$ & $65.73\pm0.00$ \\
        & MLRSSC & $53.55\pm0.30$ & $46.95\pm0.27$ & $36.54\pm0.16$ & $43.13\pm0.14$ \\
        & MCGC & $68.90\pm0.00$ & $66.84\pm0.00$ & $56.53\pm0.00$ & $61.16\pm0.00$ \\
        & CSMSC & $89.79\pm0.99$ & $82.66\pm0.65$ & $79.81\pm1.37$ & $81.83\pm1.23$ \\
        & MCLES & $82.68\pm0.10$ & $88.93\pm0.19$ & $80.41\pm0.11$ & $82.51\pm0.10$ \\
        HW & LMSC & $80.33\pm3.60$ & $79.02\pm2.07$ & $71.88\pm3.38$ & $74.78\pm3.01$ \\
        & TBGL & $98.05\pm0.00$ & $96.04\pm0.00$ & $95.54\pm0.00$ & $95.98\pm0.00$ \\
        & LTBPL & ${\color{blue}{99.55\pm0.00}}$ & ${\color{blue}{98.83\pm0.00}}$ & ${\color{blue}{99.00\pm0.00}}$ & ${\color{blue}{99.10\pm0.00}}$\\
        & WTSNM & $65.65\pm0.00$ & $58.04\pm0.00$ & $50.40\pm0.00$ & $55.39\pm0.00$\\
        & MCLGF & $\underline{98.60\pm0.00}$ & $\underline{96.69\pm0.00}$ & $\underline{96.92\pm0.00}$ & $\underline{97.23\pm0.00}$\\
        & \textbf{Ours} & ${\color{red}{99.82\pm0.03}}$ & ${\color{red}{99.59\pm0.08}}$ & ${\color{red}{99.59\pm0.07}}$ & ${\color{red}{99.63\pm0.07}}$ \\
        & \textbf{Our(random)} & $88.11\pm4.39$ & $95.33\pm1.64$ & $88.80\pm3.70$ & $90.00\pm3.26$\\
        \hline
        \hline
         & $\rm{LRR_{BestSV}}$ & $22.17\pm0.61$ & $1.85\pm0.90$ & $0.18\pm0.19$ & $17.02\pm0.48$ \\
        & Co-reg & $24.50\pm0.00$ & $20.99\pm0.19$ & $2.64\pm0.01$ & $29.03\pm0.02$ \\
        & AWP & $24.33\pm0.00$ & $21.42\pm0.00$ & $2.61\pm0.00$ & $29.05\pm0.00$ \\
        & MLRSSC & $42.37\pm0.46$ & $25.77\pm0.35$ & $13.91\pm0.55$ & $31.71\pm0.12$ \\
        & MCGC & $31.67\pm0.00$ & $25.10\pm0.00$ & $4.31\pm0.00$ & $28.94\pm0.00$ \\
        & CSMSC & $50.43\pm0.41$ & $32.71\pm0.41$ & $22.88\pm0.25$ & $36.60\pm0.26$ \\
        & MCLES & $36.82\pm5.17$ & $33.13\pm4.82$ & $12.88\pm3.89$ & $34.51\pm2.03$ \\
        Reuters & LMSC & $37.03\pm1.28$ & $26.42\pm1.28$ & $12.43\pm1.19$ & $31.87\pm0.79$ \\
        & TBGL & $21.33\pm0.00$ & $12.65\pm0.00$ & $0.68\pm0.00$ & $28.22\pm0.00$ \\
        & LTBPL & ${\color{blue}{57.50\pm0.00}}$ & $\underline{49.66\pm0.00}$ & $\underline{31.79\pm0.00}$ & $\underline{47.06\pm0.00}$\\
        & WTSNM & $55.00\pm0.00$ & $15.15\pm0.00$ & $10.99\pm0.00$ & $25.76\pm0.00$\\
        & MCLGF & $50.50\pm0.00$ & $37.52\pm0.00$ & $26.79\pm0.00$ & $40.89\pm0.00$\\
        & \textbf{Ours} & ${\color{red}{57.95\pm1.32}}$ & ${\color{red}{67.11\pm1.23}}$ & ${\color{red}{48.75\pm0.78}}$ & ${\color{red}{59.25\pm0.61}}$ \\
        & \textbf{Ours(random)} & $\underline{57.28\pm0.44}$ & ${\color{blue}{66.23\pm0.56}}$ & ${\color{blue}{48.25\pm0.57}}$ & ${\color{blue}{58.86\pm0.45}}$\\
        \bottomrule
    \end{tabular}
    }
    \label{ORL_MSRCV1_HW}
\end{table*}

\begin{table*}[h!]
    \centering
    \tiny
    \caption{Evaluation metrics of multi-view clustering methods on Reuters, Yale and BBCSport $(\%)$. Red font indicates the highest performance, blue font indicates the second best performance, and underline font indicates the third best performance.}
    \label{tab02}
    \resizebox{\linewidth}{!}{
    \begin{tabular}{c c c c c c}
         \bottomrule
         Datasets & Method & ACC & NMI & AR & F1 \\
        \hline
        & $\rm{LRR_{BestSV}}$ & $64.06\pm0.91$ & $66.16\pm0.55$ & $38.64\pm0.91$ & $42.84\pm0.81$ \\
        & Co-reg & $62.85\pm0.76$ & $67.16\pm0.69$ & $46.90\pm1.16$ & $50.26\pm1.09$ \\
        & AWP & $63.64\pm0.00$ & $67.96\pm0.00$ & $48.51\pm0.00$ & $51.74\pm0.00$ \\
        & MLRSSC & $67.21\pm2.13$ & $70.25\pm2.19$ & $49.38\pm3.08$ & $52.60\pm2.28$ \\
        & MCGC & $62.42\pm0.00$ & $65.17\pm0.00$ & $44.09\pm0.00$ & $47.67\pm0.00$ \\
        & CSMSC & $71.94\pm2.37$ & $72.95\pm1.58$ & $55.39\pm2.57$ & $58.22\pm2.39$ \\
        & MCLES & $69.76\pm1.35$ & $72.90\pm1.85$ & $50.05\pm3.86$ & $53.41\pm3.52$ \\
        Yale & LMSC & ${\color{blue}{74.30\pm1.49}}$ & $76.81\pm0.75$ & ${\color{blue}{59.00\pm1.14}}$ & ${\color{blue}{61.63\pm1.06}}$ \\
        & TBGL & $69.09\pm0.00$ & $72.53\pm0.00$ & $50.06\pm0.00$ & $53.47\pm0.00$ \\
        & LTBPL & $69.39\pm0.00$ & $71.97\pm0.00$ & $29.93\pm0.00$ & $35.61\pm0.00$\\
        & WTSNM & $73.33\pm0.00$ & $71.86\pm0.00$ & $51.01\pm0.00$ & $54.16\pm0.00$\\
        & MCLGF & $\underline{73.94\pm0.00}$ & ${\color{blue}{77.06\pm0.00}}$ & ${\color{red}{61.09\pm0.00}}$ & ${\color{red}{63.54\pm0.00}}$\\
        & \textbf{Ours} & ${\color{red}{76.12\pm3.45}}$ & ${\color{red}{78.02\pm2.19}}$ & $\underline{58.34\pm3.62}$ & $\underline{61.01\pm3.38}$ \\
        & \textbf{Ours(random)} & $73.33\pm2.73$ & $\underline{76.96\pm1.55}$ & $55.91\pm3.40$ & $58.77\pm3.12$\\
        \hline
        \hline
        & $\rm{LRR_{BestSV}}$ & $60.86\pm0.06$ & $59.16\pm0.20$ & $39.26\pm0.04$ & $59.06\pm0.02$ \\
        & Co-reg & $60.29\pm0.39$ & $47.96\pm0.72$ & $34.32\pm0.26$ & $50.68\pm0.18$ \\
        & AWP & $63.42\pm0.00$ & $50.07\pm0.00$ & $37.99\pm0.00$ & $56.36\pm0.00$ \\
        & MLRSSC & $90.61\pm0.06$ & $78.02\pm0.13$ & $81.64\pm0.12$ & $85.95\pm0.09$ \\
        & MCGC & $\underline{97.43\pm0.00}$ & $\underline{91.45\pm0.00}$ & $\underline{93.76\pm0.00}$ & $\underline{95.25\pm0.00}$ \\
        & CSMSC & $95.59\pm0.00$ & $86.19\pm0.00$ & $88.82\pm0.00$ & $91.48\pm0.00$ \\
        & MCLES & $87.98\pm0.31$ & $82.93\pm1.28$ & $84.12\pm1.02$ & $88.05\pm0.76$ \\
        BBCSport & LMSC & $86.47\pm10.73$ & $75.70\pm13.03$ & $76.77\pm17.72$ & $82.61\pm12.71$ \\
        & TBGL & $52.02\pm0.00$ & $25.08\pm0.00$ & $13.40\pm0.00$ & $42.76\pm0.00$ \\
        & LTBPL & $86.21\pm0.00$ & $79.78\pm0.00$ & $79.60\pm0.00$ & $84.78\pm0.00$\\
        & WTSNM & $50.74\pm0.00$ & $23.48\pm0.00$ & $16.72\pm0.00$ & $34.86\pm0.00$\\
        & MCLGF & $84.38\pm0.00$ & $68.95\pm0.00$ & $66.90\pm0.00$ & $75.15\pm0.00$\\
        & \textbf{Ours} & ${\color{red}{99.60\pm0.12}}$ & ${\color{red}{98.49\pm0.40}}$ & ${\color{red}{98.94\pm0.35}}$ & ${\color{red}{99.20\pm0.27}}$ \\
        & \textbf{Ours(random)} & ${\color{blue}{99.52\pm0.13}}$ & ${\color{blue}{98.21\pm0.48}}$ & ${\color{blue}{98.79\pm0.32}}$ & ${\color{blue}{99.08\pm0.24}}$\\
        \hline
        \hline
         & $\rm{LRR_{BestSV}}$ & $32.76\pm0.44$ & $16.66\pm0.72$ & $5.81\pm0.38$ & $31.67\pm0.43$ \\
        & Co-reg & $21.06\pm0.00$ & $12.17\pm0.00$ & $18.46\pm0.00$ & $26.37\pm0.00$ \\
        & AWP & $34.17\pm0.00$ & $29.27\pm0.00$ & $32.38\pm0.00$ & $39.76\pm0.00$ \\
        & MLRSSC & $45.96\pm1.21$ & $25.01\pm0.98$ & $20.97\pm0.56$ & $36.80\pm1.15$ \\
        & MCGC & $54.19\pm0.00$ & $\underline{64.48\pm0.00}$ & $51.72\pm0.00$ & $59.42\pm0.00$ \\
        & CSMSC & $58.36\pm0.26$ & $47.25\pm0.86$ & $44.45\pm0.25$ & $54.47\pm0.00$ \\
        & MCLES & $61.15\pm2.14$ & $54.98\pm1.97$ & $\underline{58.34\pm3.46}$ & $\underline{65.24\pm2.16}$ \\
        BDGP & LMSC & $47.24\pm0.05$ & $24.22\pm0.03$ & $31.04\pm0.02$ & $35.48\pm0.08$ \\
        & TBGL & $57.80\pm0.00$ & $54.65\pm0.00$ & $50.16\pm0.00$ & $55.64\pm0.00$ \\
		& LTBPL & $38.32\pm0.00$ & $16.45\pm0.00$ & $35.96\pm0.00$ & $32.81\pm0.00$\\
		& WTSNM & $32.04\pm0.00$ & $6.69\pm0.00$ & $18.16\pm0.00$ & $27.27\pm0.00$\\ 
		& MCLGF & $\underline{62.40\pm0.00}$ & $38.16\pm0.00$ & $35.31\pm0.00$ & $50.46\pm0.00$\\    
        & \textbf{Ours} & ${\color{red}{93.48\pm1.67}}$ & ${\color{red}{88.00\pm1.56}}$ & ${\color{red}{85.81\pm2.95}}$ & ${\color{red}{88.67\pm2.34}}$ \\
        & \textbf{Ours(random)} & ${\color{blue}{81.65\pm1.96}}$ & ${\color{blue}{79.25\pm2.48}}$ & ${\color{blue}{71.97\pm2.11}}$ & ${\color{blue}{77.94\pm1.51}}$\\
         \hline
        \hline
        & $\rm{LRR_{BestSV}}$ & $29.68\pm0.03$ & $31.28\pm0.06$ & $14.61\pm0.11$ & $22.05\pm0.02$ \\
        & Co-reg & $41.05\pm0.00$ & $35.14\pm0.00$ & $46.18\pm0.00$ & $39.17\pm0.00$ \\
        & AWP & $38.18\pm0.00$ & $29.35\pm0.00$ & $34.67\pm0.00$ & $39.76\pm0.00$ \\
        & MLRSSC & $37.26\pm1.33$ & $37.68\pm1.04$ & $21.90\pm0.78$ & $27.33\pm1.26$ \\
        & MCGC & $59.21\pm0.00$ & $63.69\pm0.00$ & $53.17\pm0.00$ & $57.73\pm0.00$ \\
        & CSMSC & $64.24\pm2.64$ & $68.19\pm1.68$ & $59.38\pm2.43$ & $63.17\pm1.75$ \\
        & MCLES & $68.16\pm1.13$ & $72.64\pm2.15$ & $64.13\pm1.86$ & $69.38\pm2.64$ \\
        Scene15 & LMSC & $60.28\pm1.39$ & $68.46\pm0.79$ & $59.64\pm1.23$ & $62.94\pm1.06$ \\
        & TBGL & $78.45\pm0.00$ & $69.34\pm0.00$ & $75.58\pm0.00$ & $81.15\pm0.00$ \\
		& LTBPL & $79.58\pm0.00$ & $65.83\pm0.00$ & $72.76\pm0.00$ & $\underline{82.33\pm0.00}$\\
		& WTSNM & $72.46\pm0.00$ & $58.68\pm0.00$ & $71.05\pm0.00$ & $76.12\pm0.00$\\
		& MCLGF & $\underline{81.59\pm0.00}$ & $\underline{82.46\pm0.00}$ & $\underline{79.14\pm0.00}$ & ${\color{red}{83.08\pm0.00}}$\\		       
        & \textbf{Ours} & ${\color{red}{84.16\pm2.74}}$ & ${\color{red}{89.38\pm0.51}}$ & ${\color{red}{81.49\pm1.58}}$ & ${\color{blue}{82.83\pm1.44}}$ \\
        & \textbf{Ours(random)} & ${\color{blue}{82.97\pm1.81}}$ & ${\color{blue}{89.01\pm0.00}}$ & ${\color{blue}{80.67\pm0.72}}$ & $82.08\pm0.65$\\
        \bottomrule
    \end{tabular}
    }
    \label{Reuters_Yale_BBCSport}
\end{table*}

\begin{table*}[h!]
    \centering
    \tiny
    \caption{Ablation experiments on different datasets $(\%)$. Red font indicates the highest performance. }
    \label{tab_r1}
    \resizebox{\linewidth}{!}{
    \begin{tabular}{c c c c c c c c c c c c}
         \bottomrule
         Datasets & Methods & ACC & NMI & AR & F1 & Datasets & Methods & ACC & NMI & AR & F1\\
        \hline
        & ELMSC & ${\color{red}{90.50}}$ & ${\color{red}{96.43}}$ & ${\color{red}{88.31}}$ & ${\color{red}{88.59}}$ & & ELMSC & ${\color{red}{98.57}}$ & ${\color{red}{97.03}}$ & ${\color{red}{96.68}}$ & ${\color{red}{97.14}}$\\
        ORL & ELMSC-v1 & $86.50$ & $92.54$ & $80.17$ & $80.63$ & MSRCV1 & ELMSC-v1 & $92.38$ & $89.87$ & $84.65$ & $86.81$\\
        & ELMSC-v2 & $83.75$ & $91.98$ & $76.61$ & $77.16$ & & ELMSC-v2 & $59.52$ & $56.94$ & $44.31$ & $52.69$\\
        \hline
		& ELMSC & ${\color{red}{99.82}}$ & ${\color{red}{99.59}}$ & ${\color{red}{99.59}}$ & ${\color{red}{99.63}}$ & & ELMSC & ${\color{red}{57.95}}$ & ${\color{red}{67.11}}$ & ${\color{red}{48.75}}$ & ${\color{red}{59.25}}$\\
		HW & ELMSC-v1 & $97.75$ & $97.39$ & $97.45$ & $97.50$ & Reuters & ELMSC-v1 & $54.17$ & $58.44$ & $42.10$ & $52.83$\\
		& ELMSC-v2 & $80.80$ & $83.79$ & $76.75$ & $79.23$ & & ELMSC-v2 & $24.67$ & $22.35$ & $3.49$ & $29.33$\\
		\hline
        & ELMSC & ${\color{red}{76.12}}$ & ${\color{red}{78.02}}$ & ${\color{red}{58.34}}$ & ${\color{red}{61.01}}$ & & ELMSC & ${\color{red}{99.60}}$ & ${\color{red}{98.49}}$ & ${\color{red}{98.94}}$ & ${\color{red}{99.20}}$\\
		Yale & ELMSC-v1 & $73.55$ & $76.16$ & $59.58$ & $62.03$ & BBCSport & ELMSC-v1 & $97.08$ & $94.88$ & $95.74$ & $96.27$\\
		& ELMSC-v2 & $47.27$ & $51.07$ & $21.57$ & $26.80$ & & ELMSC-v2 & $94.67$ & $84.32$ & $85.64$ & $89.01$\\        		        
        \bottomrule
    \end{tabular}
    }
    \label{Ablation}
\end{table*}  

\begin{table*}[h!]
    \centering
    \tiny
    \caption{Comparison results of tensor-based methods and our method with random parameter selection on different datasets $(\%)$. Red font indicates the highest performance. }
    \label{tab_r2}
    \resizebox{\linewidth}{!}{
    \begin{tabular}{c c c c c c c c c c c c}
         \bottomrule
         Datasets & Methods & ACC & NMI & AR & F1 & Datasets & Methods & ACC & NMI & AR & F1\\
        \hline
        \multirow{4}{*}{ORL} & TBGL & $53.17$ & $64.66$ & $42.37$ & $55.34$ & \multirow{4}{*}{MSRCV1} & TBGL & $80.94$ & $74.13$ & $79.49$ & $74.27$\\
        & LTBPL & $45.75$ & $69.84$ & $9.86$ & $13.60$ &  & LTBPL & $15.24$ & $5.35$ & $0.05$ & $24.23$\\
        & WTSNM & $83.75$ & $91.31$ & $76.62$ & $77.16$ & & WTSNM & $79.06$ & $66.66$ & $58.14$ & $64.07$\\
		& \textbf{Ours} & ${\color{red}{85.88}}$ & ${\color{red}{95.53}}$ & ${\color{red}{83.54}}$ & ${\color{red}{83.93}}$ & & \textbf{Ours} & ${\color{red}{98.43}}$ & ${\color{red}{96.75}}$ & ${\color{red}{96.33}}$ & ${\color{red}{96.84}}$\\        
        \hline
		\multirow{4}{*}{HW} & TBGL & $64.19$ & $54.27$ & $49.76$ & $59.14$ & \multirow{4}{*}{Reuters} & TBGL & $28.21$ & $9.46$ & $12.06$ & $34.94$\\
		& LTBPL & $13.80$ & $7.94$ & $0.55$ & $18.42$ &  & LTBPL & $17.17$ & $1.62$ & $0.00$ & $28.30$\\
		& WTSNM & $65.60$ & $57.98$ & $50.33$ & $55.33$ & & WTSNM & $30.00$ & $11.83$ & $5.50$ & $25.21$\\
		& \textbf{Ours} & ${\color{red}{88.11}}$ & ${\color{red}{95.33}}$ & ${\color{red}{88.80}}$ & ${\color{red}{90.00}}$ & & \textbf{Ours} & ${\color{red}{57.28}}$ & ${\color{red}{66.23}}$ & ${\color{red}{48.25}}$ & ${\color{red}{58.86}}$\\
		\hline
        \multirow{4}{*}{Yale} & TBGL & $52.37$ & $50.73$ & $37.19$ & $42.61$ &  \multirow{4}{*}{BBCSport} & TBGL & $24.19$ & $4.78$ & $5.34$ & $20.53$\\
		& LTBPL & $30.30$ & $42.05$ & $5.21$ & $15.59$ & & LTBPL & $38.24$ & $9.49$ & $2.47$ & $39.49$\\
		& WTSNM & $73.33$ & $71.86$ & $51.01$ & $54.16$ & & WTSNM & $37.68$ & $23.74$ & $6.51$ & $32.45$\\
		& \textbf{Ours} & ${\color{red}{73.33}}$ & ${\color{red}{76.96}}$ & ${\color{red}{55.91}}$ & ${\color{red}{58.77}}$ & & \textbf{Ours} & ${\color{red}{99.52}}$ & ${\color{red}{98.21}}$ & ${\color{red}{98.79}}$ & ${\color{red}{99.08}}$\\
		\hline
		\multirow{4}{*}{BDGP} & TBGL & $19.76$ & $2.43$ & $1.79$ & $18.37$ & \multirow{4}{*}{Scene15} & TBGL & $62.76$ & $68.19$ & $59.43$ & $65.24$\\ 
		& LTBPL & $38.24$ & $9.49$ & $2.47$ & $39.49$ & & LTBPL & $51.18$ & $59.72$ & $43.37$ & $55.49$\\
		& WTSNM & $25.36$ & $3.18$ & $2.49$ & $24.52$ & & WTSNM & $42.36$ & $25.71$ & $19.64$ & $36.38$\\
		& \textbf{Ours} & ${\color{red}{81.65}}$ & ${\color{red}{79.25}}$ & ${\color{red}{71.97}}$ & ${\color{red}{77.94}}$ & & \textbf{Ours} & ${\color{red}{82.97}}$ & ${\color{red}{89.01}}$ & ${\color{red}{80.67}}$	& ${\color{red}{82.08}}$\\	 		        
        \bottomrule
    \end{tabular}
    }
    \label{tenosr_random}
\end{table*}



In this section, we conduct comprehensive experiments to evaluate the effectiveness of the proposed ELMSC algorithm. Specifically, the clustering performance compared with other methods, affinity matrix and convergence analysis, parameter sensitivity analysis, and data visualization are conducted to perform the validation.  Experiments are conducted on a Windows 11 PC with a Intel i5-12490F processor at 3.00 GHz and 48GB RAM, using MATLAB R2023a for all analyses. All results are obtained by averaging 10 independent trials. The source code will be released on \href{https://github.com/caolei2000/ELMSC-Code}{https://github.com/caolei2000/ELMSC-Code} soon.
%

\subsection{Experimental Settings}

\subsubsection{Datasets} We select six real-world multi-view datasets, including \textbf{ORL} \cite{luo2018consistent}, \textbf{MSRCV1} \cite{winn2005locus}, \textbf{HW} \cite{asuncion2007uci}, \textbf{Reuters} \cite{apte1994automated}, \textbf{Yale} \cite{wang2021multi, xia2021multiview} and \textbf{BBCSport} \cite{chen2021low}. A brief description on these datasets is given below:

\textbf{ORL}\footnotemark \footnotetext{https://www.cl.cam.ac.uk/research/dtg/attarchive/facedatabase.html}: It comprises 400 facial images from 40 subjects, each represented by 10 images. The subject label serves as the definitive class label. Images were captured under varying conditions, including lighting, time, and facial expressions or details. Three types of features are utilized: intensity, LBP, and Gabor. 

\textbf{MSRCV1}\footnotemark \footnotetext{http://research.microsoft.com/en-us/projects/objectclassrecognition}: It is composed of 210 images for scene recognition, distributed across seven categories. Each image is characterized by six different feature sets: LBP with 256 dimensions, HOG with 100 dimensions, GIST with 512 dimensions, Color Moment with 48 dimensions, CENTRIST with 1302 dimensions, and SIFT with 210 dimensions.

\textbf{HW}\footnotemark \footnotetext{https://cs.nyu.edu/~roweis/data.html}: It is composed of 2,000 data points for digits 0 to 9 from UCI machine learning repository and two public features are available. 

\textbf{Reuters}\footnotemark \footnotetext{http://lig-membres.imag.fr/grimal/data.html}: It is a dataset of newswire articles available in five languages: French, Italian, English, German, and Spanish. Our experiment is conducted on a subset comprising 600 documents from 6 articles. 

\textbf{Yale}\footnotemark \footnotetext{http://cvc.yale.edu/projects/yalefacesB/yalefacesB.html}: It consists of 165 gray-scale images of 15 individuals with different facial expressions and configurations. Motivated by \cite{chen2022low}, 4096 dimensions intensity feature, 3304 LBP feature and 6750 dimensions Gabor feature are extracted as three multi-view features.

\textbf{BBCSport}\footnotemark \footnotetext{http://mlg.ucd.ie/datasets/bbc.html}: It comprises 544 documents sourced from the BBC Sport website, covering news from five different categories - business, entertainment, politics, sport, and tech. These categories serve as the ground-truth class labels. In the experiments, two views with dimensions of 3183 and 3203 are selected. 

\text{BDGP}\footnotemark \footnotetext{https://www.fruitfly.org/}: It contains 2,500 images about drosophila embryos belonging to 5 categories. Each image is represented by a 1,750-D visual vector and a 79-D textual feature vector.

\text{Scene15}\cite{fei2005bayesian}: It consists of 4,485 images from 15 indoor and outdoor scene categories. Three types features (GIST, PHOG and LBP) are extracted.

\subsubsection{Compared Methods}

We compare our method with some baseline methods including: $\mathbf{LRR_{BestSV}}$ \cite{liu2012robust}, \textbf{Co-reg} \cite{kumar2011co}, \textbf{AWP} \cite{nie2018multiview}, \textbf{MLRSSC} \cite{brbic2018multi}, \textbf{MCGC} \cite{zhan2018multiview}, \textbf{CSMSC} \cite{luo2018consistent}, \textbf{MCLES} \cite{chen2020multi},  \textbf{LMSC} \cite{zhang2018generalized}, \textbf{TBGL} \cite{xia2022tensorized}, \textbf{LTBPL} \cite{chen2022low}, \textbf{WTSNM} \cite{xia2021multiview} and \textbf{MCLGF} \cite{zhou2023learnable}. The compared methods are detailed below:

$\mathbf{LRR_{BestSV}}$ : This method employs the nuclear norm to construct the representation matrix, a technique used in single-view clustering.  

\textbf{Co-reg}: This method constructs an optimal consensus adjacency matrix and uses the nuclear norm to project data into a low-dimensional subspace. 

\textbf{AWP}: This method introduces an adaptively weighted procrustes strategy that weighs views based on their clustering capacities. 

\textbf{MLRSSC}: This method incorporates sparsity and low-rankness to learn the representation matrix.

\textbf{MCGC}: This method learns a consensus graph by minimizing disagreement between different views and constraining the rank of the Laplacian matrix.

\textbf{CSMSC}: This method takes into account consistency and specificity jointly for subspace representation learning.

\textbf{MCLES}: This method clusters multi-view data in a learned latent embedding space.

\textbf{LMSC}: This method proposes a latent representation scheme to exploit comprehensive information from multiple views.

\textbf{TBGL}: This method uses a variance-based de-correlation anchor selection strategy for bipartite construction.

\textbf{LTBPL}: This method proposes a framework to simultaneously consider the low-rank probability affinity matrices and the integrated consensus indicator graph.

\textbf{WTSNM}: This method investigates the weighted tensor Schatten \emph{p}-norm to exploit the significance of different singular values in tensor-singular value decomposition. 

\textbf{MCLGF}: This method constructs a consensus graph filter by considering the information across different views. 

\subsubsection{Evaluation Metrics and Parameter Settings}

We evaluate the performance of all compared multi-view clustering algorithms using four widely used metrics:  clustering accuracy (ACC), normalized mutual information (NMI), adjusted rand index (AR), and F1 score. For a fair comparison, the associated regularization parameters of all methods are carefully tuned over the range $[10^{-3}, 10^3]$ with a grid size $\{10^{-3}, 10^{-2}, 10^{-1}, 10^0, 10^1, 10^2, 10^3\}$ to achieve optimal performance. This approach to parameter selection may differ from that of existing methods; thus it is expected that our optimal results for these methods may deviate from those reported in the corresponding literature.

\begin{figure*}[htbp]
	\centering
	\includegraphics[scale=0.9]{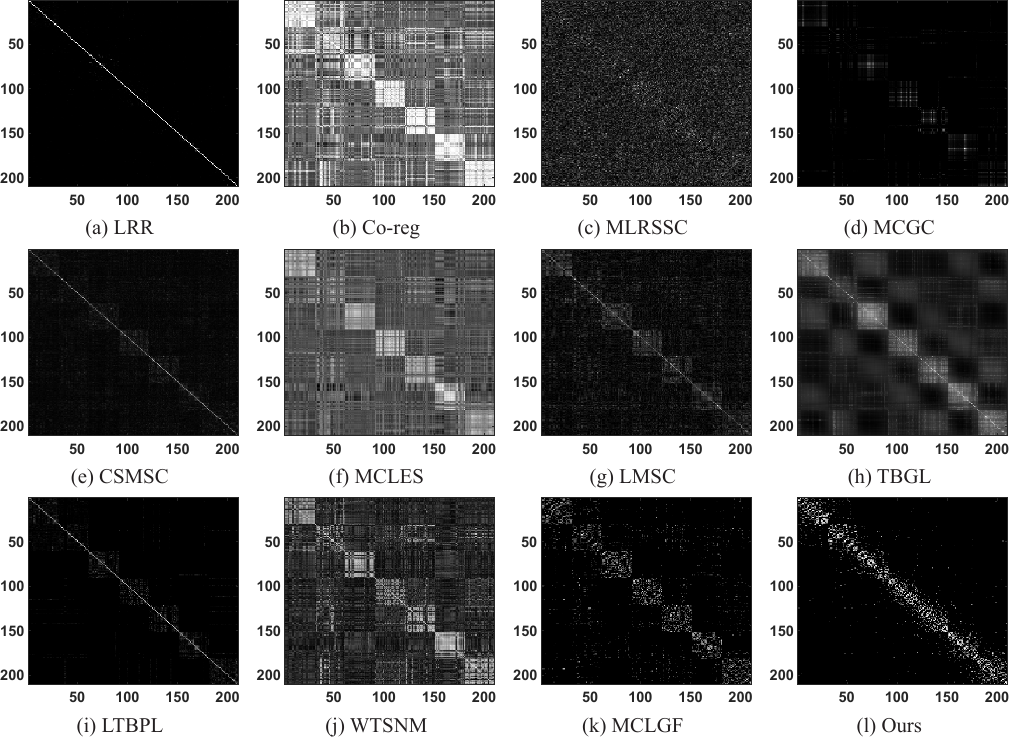} 
	\caption{Visualization of affinity matrix on MSRCV1.}
	\label{Fig_aff}
\end{figure*}

\begin{figure}[htbp]
	\centering
	\includegraphics[scale=0.9]{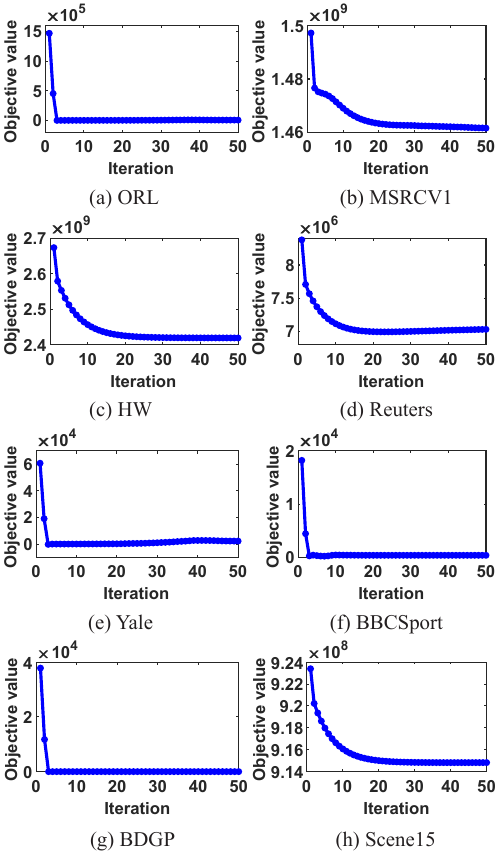}
	\caption{Convergence of ELMSC.}
    \label{Fig_cov}
\end{figure}

\begin{figure*}[htbp]
\centering
\includegraphics[scale=0.85]{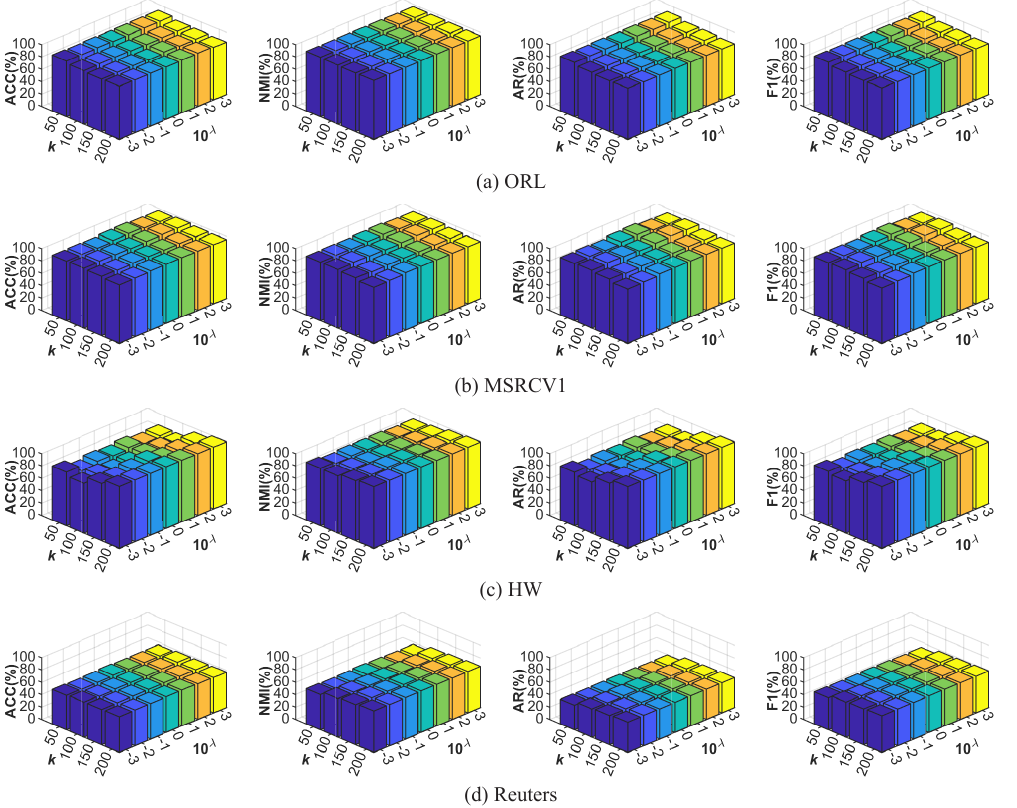} 
\caption{Parameter sensitivity analysis of the proposed method.}
\label{Fig_sen}
\end{figure*}

\begin{figure*}[htbp]
\centering
\includegraphics[scale=0.85]{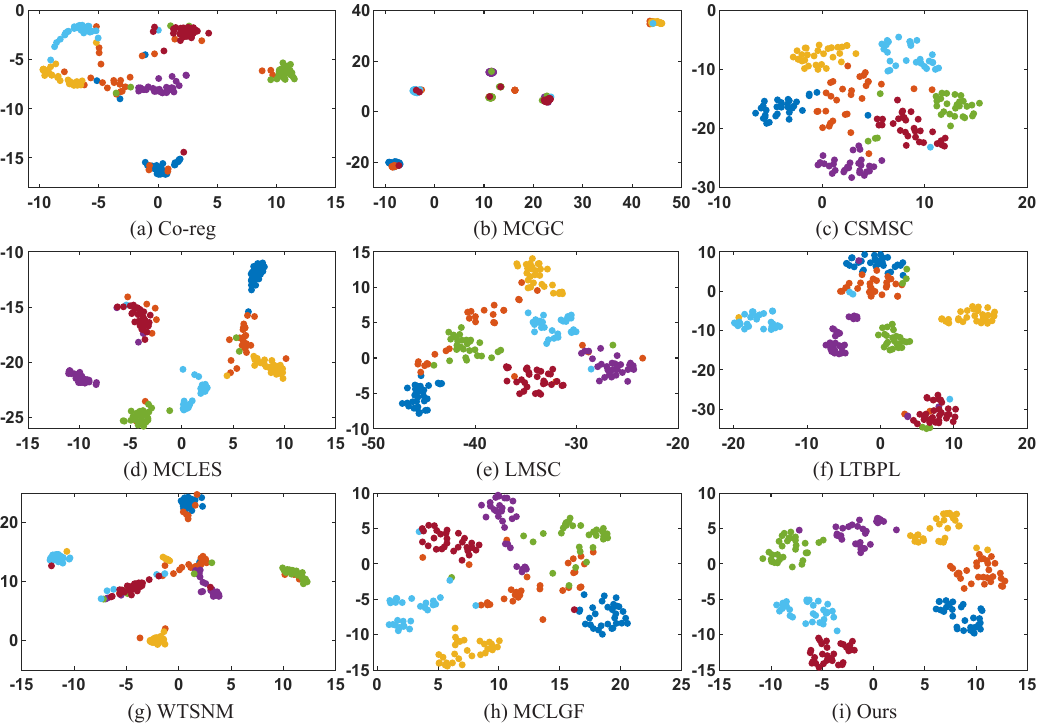} 
\caption{T-SNE visualization of various methods on MSRCV1.}
\label{Fig_tsne}
\end{figure*}

\begin{figure*}[htbp]
	\centering
	\includegraphics[scale=0.85]{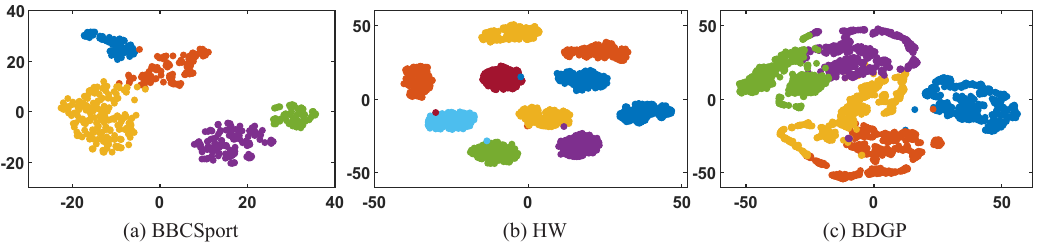}
	\hspace{2cm}\caption{T-SNE visualization results of our method on different datasets.} 
	\label{Fig_r2}
\end{figure*}

\begin{figure*}[htbp]
	\centering
	\includegraphics[scale=0.85]{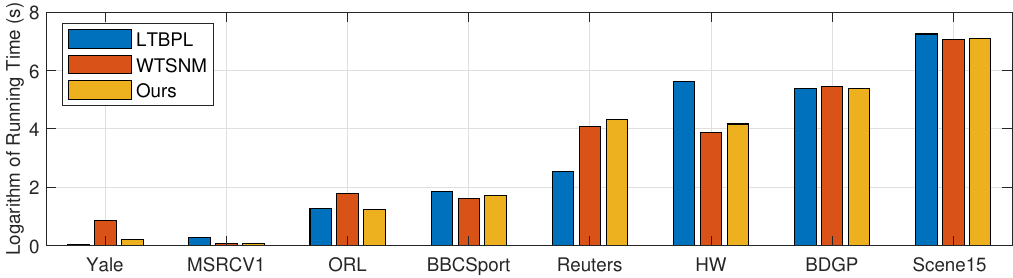}
	\hspace{2cm}\caption{Actual running time between LTBPL, WTSNM and our method on different datasets.} 
	\label{Fig_r1}
\end{figure*}

\subsubsection{Implementation of PCA}

As mentioned below Eq. (\ref{eq06}), our method needs to perform PCA for data samples to align dimensions. For implementing PCA, we reduce the data dimensionality across various views to a consistent $d$ $(d<\min(d_i,d_j))$. The determination of the number of principal components is influenced by the number of clusters. More specifically, we set the number of principal components to be $clusters\times 6$. Generally, $clusters\times 6$ does not exceed the number of samples $n$. Thus, the number of principal components meets the requirement of being within the range of $0$ and $\min(n, d_i)$. In the implementation of PCA, the amount of information is typically measured by
the variance of the data. The percentage of the original information is calculated by dividing the sum of variances of the selected principal components by the total variance of all components. We have tested that our PCA implementation remains about $92\%\sim98\%$ information. 


\subsection{Experimental Results}

\subsubsection{Clustering Analysis}

Tables \ref{ORL_MSRCV1_HW} and \ref{Reuters_Yale_BBCSport} record the clustering results of all involved multi-view clustering methods on different datasets. In the tables, we use red font to indicate the highest performance, blue font to indicate the second best performance, and underline font to indicate the third best performance. Note that we include two sets of results from our method in the tables, labeled as ``{\bf{Ours}}" and ``{\bf{Ours(random)}}", where the results for ``{\bf{Ours}}" are obtained by selecting optimal regularization parameter according to the previously mentioned parameter settings, while the results for ``{\bf{Ours(random)}}" are obtained by randomly selecting regularization parameter from the gird set $\{10^{-3}, 10^{-2}, 10^{-1}, 10^0, 10^1, 10^2, 10^3\}$. Beside choosing the regularization parameter $\lambda$, our method also involves $k$, representing the dimensionality of latent representation, is optimally selected from the grid set $[50,100,150,200]$ for ``{\bf{Ours}}", and randomly selected from the same set for ``{\bf{Ours(random)}}". 

It is seen that $\mathrm{LRR_{BestSV}}$, a representative single-view clustering method, is generally inferior to all other multi-view methods in the comparison. This demonstrates the advantages of multi-view methods, which are capable of exploiting comprehensive information. We also observe that among previous methods, tensor-based methods, namely TBGL, LTBPL and WTSNM, generally provide more superior clustering performance. This suggests that tensor-based methods possess a more potent capability for data representation. 

In all experiments conducted across various datasets, our proposed ELMSC algorithm with optimal parameter selection consistently shows remarkable advancement over the competing baselines in terms of all performance metrics. The only exception is the AR metric, where it ranks third on the Yale dataset. For instance, on the ORL dataset, our method with optimal parameter selection yields an improvement of $6.25\%$ in terms of ACC when contrasted with WTSNM which achieves the highest performance among previous methods. Similarly, it shows an improvement of $2.71\%$ in terms of NMI when compared to LTBPL, another top-performing method from previous studies. This is because the data construction method in the ELMSC significantly enhances the representation capability, thereby facilitating a more comprehensive recovery of the latent space.  

Additionally, it is seen that our method with random parameter selection shows performance that is second only to that of our method with optimal parameter selection on the ORL, MSRCV1, Reuters, BDGP, Scene15 datasets. It outperforms all other methods in terms of ACC and NMI. For example, on the MSRCV1 dataset, when compared to TBGL, which achieves the highest performance among previous methods, our method with random parameter selection shows improvements of  $0.33\%$ and $0.72\%$ in terms of ACC and NMI, respectively. Even on other datasets, our method with random parameter selection yields performance that is comparable to the best-performing method among previous ones. For instance, on the Reuters dataset, our method with random parameter selection achieves an ACC $57.28\%$ that is very close to the $57.50\%$ achieved by the top-performing previous method, LTBPL. This illustrates that our method’s performance is robust to parameter selection, demonstrating its reliability and generalizability in real-world scenarios. This advantage will be further verified in the forthcoming visualization of the parameter sensitivity analysis.

\subsubsection{Ablation studies}

We perform ablation experiments to investigate the contributions of the sparse regularization on non-diagonal blocks and the augmented data construction strategy. Specifically, for the first test ($\lambda=0$), the term of sparse regularization of non-diagonal blocks is ignored. And for the second test, we remain the diagonal blocks unchanged but replace non-diagonal blocks with zero matrices, effectively omitting the inter-view similarity information. For notation simplicity, we denote the first test as ELMSC-v1, and the second test as ELMSC-v2. The comparison results are shown in Table \ref{tab_r1}.

It is observed that ELMSC achieves more advanced performance compared with ELMSC-v1 and ELMSC-v2 across all test cases, which demonstrates that each component contributes to the overall performance. Notably, we observe that ELMSC-v2 exhibits a remarkable decrease compared to ELMSC, particularly on the Yale and Reuters datasets. This implies that the augmented data construction strategy plays a more significant role in enhancing algorithm performance.

\subsubsection{Comparison with tensor-based methods}

Our method requires tuning only one hyperparameter, $\lambda$, as the dimensionality of the latent representation, $k$, is typically set to $100$ \cite{zhang2018generalized}. In contrast, many tensor-based clustering methods require three or more hyperparameters. To further analyze the advantages of our method compared to tensor-based methods, we perform experiments under random parameter selection. For tensor-based methods, their parameters are randomly selected from $\{10^{-3}, 10^{-2}, \cdots, 10^2, 10^3\}$. The comparison results on different datasets are shown in Table \ref{tab_r2}. It is seen that under the condition of random parameter selection, our method significantly outperforms TBGL, LTBPL, and WTSNM, particularly on the BBCSport, BDGP, and Scene15 datasets.

\subsubsection{Affinity Matrix and Convergence Analysis}

Fig. \ref{Fig_aff} shows the visualization results of the affinity matrix of various methods on MSRCV1. Compared with other methods, we can observe a clearer block-diagonal structure in the affinity matrix when using our method. It implies that the data samples can be effectively partitioned into distinct clusters. Fig. \ref{Fig_cov} shows the evolutionary curves of the objective function value of ELMSC on the ORL, MSRCV1, HW, Reuters, Yale, BBCSport, BDGP, and Scene15 datasets, which empirically illustrates that the iterative algorithm in ELMSC is convergent. 

\subsubsection{Parameter Sensitivity Analysis}

For conducting the parameter sensitivity analysis, we investigate the performance of our method with respect to the regularization parameter $\lambda$ and the dimensionality of latent representation $k$. Fig. \ref{Fig_sen} shows the visualization results of parameter sensitivity on four datasets. Our method consistently achieves stable performance for different parameters in terms of all performance metrics. An exception is observed with the HW dataset, where our proposed method experiences minor fluctuations with different choices of $k$. Overall, our method's performance is not significantly sensitive to parameter selection. Therefore, even when our method operates with randomly selected parameters, it is expected to yield a satisfactory performance that closely approximates the optimal performance. 

\subsubsection{Visualization of Clustering Results}

In order to visualize more intuitively, we present the t-Distributed Stochastic Neighbor Embedding (T-SNE) \cite{van2008visualizing} visualization results of nine methods, as show in Fig. \ref{Fig_tsne}. It is observed that tensor-based multi-view methods typically yield superior cluster partitions compared to other techniques. For instance, LTBPL demonstrates a distinct partition result, whereas CSMSC presents little messy partitions. Furthermore, in contrast with the previous methods, our method achieves the distinction of clusters in a more discriminative and clear manner.  

Furthermore, we conduct additional experiments to verify the visualization performance of ELMSC. Fig. \ref{Fig_r2} records the corresponding T-SNE results on three different datasets. We also observe clear clusters on the tested datasets.

\subsubsection{Actual running time}

Consider that our method's computational complexity is comparable to tensor-based methods, we only compare the actual running time of our method with that of tenor-based methods. In addition, since ELMSC does not utilize the anchor technique to accelerate processing, we do not compare it with anchor-guided tensor multi-view methods. The comparison results on actual running time is shown in Fig. \ref{Fig_r1}. It is observed that in comparison to tensor-based methods, our method’s running time is comparable across most datasets.




\section{Conclusions}

In this paper, we propose a novel latent multi-view subspace clustering method, namely ELMSC. In ELMSC, we construct an augmented multi-view data matrix, in which the diagonal blocks stack data matrices from each view for preserving complementary information, while the non-diagonal blocks comprise of the similarity information between different views to ensure consistency. This data construction strategy significantly contributes to a more comprehensive recovery of latent representations. In addition, we enforce a sparse regularization for the non-diagonal blocks of the augmented self-representation matrix, mitigating redundant computations of consistent information. Meanwhile, we devise a novel iterative algorithm using ADMM for ELMSC. Furthermore, we establish the relationship between LMSC and our method, and theoretically analyze the convergence of ELMSC. Experiment results on real-world datasets validate the advantages of ELMSC over several baselines. Importantly, our experiments highlight the effectiveness of our method even when parameters are selected randomly, implying the potential applicability of ELMSC in practical scenarios.  

%

\ifCLASSOPTIONcaptionsoff
  \newpage
\fi

\bibliographystyle{IEEEtran}
\bibliography{IEEEabrv,IEEEexample}


%
%
%
%
%
%
%
%

%
%
\begin{IEEEbiography}[{\includegraphics[width=1in,height=1.25in,clip,keepaspectratio]{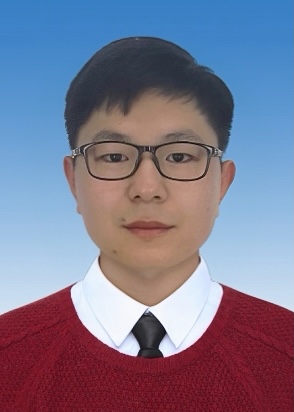}}]{Long Shi} (Member, IEEE) was born in Jiangsu Province, China. He received the Ph.D degree in electrical engineering from Southwest Jiaotong University, Chengdu, China, 2020. From 2018 to 2019, he was a visiting student with the Department of Electronic Engineering, University of York, U.K. Since 2021, he has been with the School of Computing and Artificial Intelligence, Southwestern University of Finance and Economics, where he is currently an assistant professor. He has published several high-quality papers in IEEE TSP, IEEE SPL, IEEE TCSII, etc. His research interest lies at the intersection of signal processing and machine learning, including adaptive signal processing, multi-view (multimodal) learning, large language model.\end{IEEEbiography}

\begin{IEEEbiography}[{\includegraphics[width=1in,height=1.25in,clip,keepaspectratio]{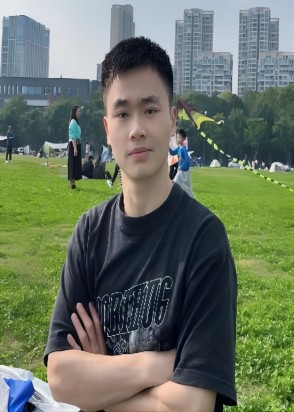}}]{Lei Cao} is currently pursuing a Master’s degree at School of Computing and Artificial Intelligence, Financial Intelligence and Financial Engineering Key Laboratory of Sichuan Province (FIFE), Southwestern University of Finance and Economics(SWUFE), supervised by Dr. Long Shi. His research interests are mainly in multi-view clustering, multimodal learning and applications in finance.\end{IEEEbiography}

\begin{IEEEbiography}[{\includegraphics[width=1in,height=1.25in,clip,keepaspectratio]{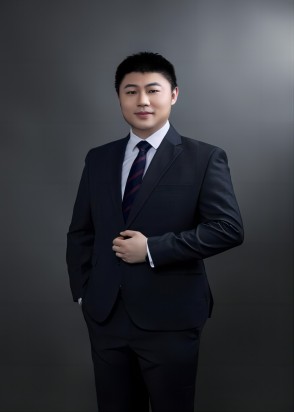}}]{Jun Wang} is currently a Professor with the School of Management Science and Engineering at the Southwestern University of Finance and Economics, where he serves as the head of the Department of Information Management. He was a visiting scholar at the Memorial University of Newfoundland, Canada. He is the Deputy Director of the Provincial Key Laboratory of Financial Intelligence and Financial Engineering, as well as the Deputy Director of the NLP Professional Committee of the Sichuan Computer Society. His research interests lie in the emerging interdisciplinary fields of financial technology, financial intelligence, and digital intelligence management.  \end{IEEEbiography}

\begin{IEEEbiography}[{\includegraphics[width=1in,height=1.25in,clip,keepaspectratio]{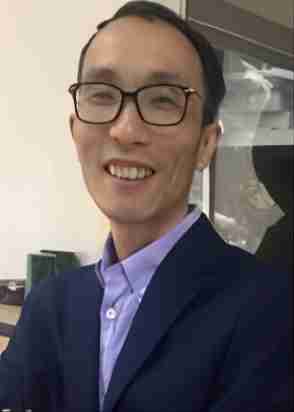}}]{Badong Chen} received the Ph.D. degree in Computer Science and Technology from Tsinghua University, Beijing, China, in 2008. He is currently a professor with the Institute of Artificial Intelligence and Robotics, Xi'an Jiaotong University, Xi'an, China. His research interests are in signal processing, machine learning, artificial intelligence and robotics. He has authored or coauthored over 300 articles in various journals and conference proceedings (with 13000+ citations in Google Scholar), and has won the 2022 Outstanding Paper Award of IEEE Transactions on Cognitive and Developmental Systems. Dr. Chen serves as a Member of the Machine Learning for Signal Processing Technical Committee of the IEEE Signal Processing Society, and serves (or has served) as an Associate Editor for several journals including IEEE Transactions on Neural Networks and Learning Systems, IEEE Transactions on Cognitive and Developmental Systems, IEEE Transactions on Circuits and Systems for Video Technology, Neural Networks and Journal of The Franklin Institute. He has served as a PC or SPC Member for prestigious conferences including UAI, IJCAI and AAAI, and served as a General Co-Chair of 2022 IEEE International Workshop on Machine Learning for Signal Processing.\end{IEEEbiography}

\end{document}